%% file: sample-sigconf.tex
\documentclass[sigconf]{acmart}

\usepackage{tikz}
\usepackage{caption}
\usepackage{filecontents}
\usepackage{url}
\usepackage{amsmath}
\usepackage{graphicx}
\usepackage{url}
\usepackage{hyperref}

\usepackage{amsfonts,amssymb}
\usepackage{color, soul}
\usepackage{mathrsfs}
\usepackage{cleveref}
\usepackage{multirow}
\usepackage{wrapfig}
\usepackage{amsthm}
\usepackage{algorithm}
\usepackage{algorithmic}
\usepackage{bm}
\usepackage{enumitem}

\newtheorem{myDef}{Definition}
\newtheorem{myDef1}{Problem}

\usepackage{tikz}
\usepackage{subcaption}
\usepackage{pgfplots}
\usetikzlibrary{arrows,positioning,automata,calc,shapes}
\pgfplotsset{compat=newest, scaled z ticks=false} 
\pgfplotsset{plot coordinates/math parser=false}
\newlength\figureheight 
\newlength\figurewidth

\usetikzlibrary{plotmarks}
\definecolor{txtcolor1}{rgb}{1,0.65,0}%

\AtBeginDocument{%
  \providecommand\BibTeX{{%
    \normalfont B\kern-0.5em{\scshape i\kern-0.25em b}\kern-0.8em\TeX}}}

\setcopyright{acmcopyright}
\copyrightyear{2022} 
\acmYear{2022} 
\setcopyright{acmcopyright}\acmConference[WWW '22]{Proceedings of the ACM Web Conference 2022}{April 25--29, 2022}{Virtual Event, Lyon, France}
\acmBooktitle{Proceedings of the ACM Web Conference 2022 (WWW '22), April 25--29, 2022, Virtual Event, Lyon, France}
\acmPrice{15.00}
\acmDOI{10.1145/3485447.3512173}
\acmISBN{978-1-4503-9096-5/22/04}

\begin{document}
% \fancyhead{}
% \renewcommand{\shortauthors}{}
% \title{EDITS: Model-Agnostic Dual Debiasing for Attributed Network Mining}

% \title{Dual Debiasing for Graph Neural Networks: \\ A Task-Agnostic Approach}
% \title{EDITS: Task-Agnostic Debiasing for Graph Neural Networks}
\title{EDITS: Modeling and Mitigating Data Bias \\ for Graph Neural Networks}
% \title{EDITS: Modeling and Mitigating Data Bias \\ for nnn Networks}
% Contrastive ?  self-supervised learning ?

% \affiliation{%

% \author{Yushun Dong$^1$, Ninghao Liu$^2$, Brian Jalaian$^3$, Jundong Li$^1$}
% \affiliation{%
%   \institution{$^1$University of Virginia, USA}
%   \institution{$^2$University of Georgia, USA}  
%   \institution{$^3$Army Research Laboratory, USA}}
% \email{{yd6eb,jundong}@virginia.edu,ninghao.liu@uga.edu,brian.a.jalaian.civ@mail.mil}

\author{Yushun Dong$^1$, Ninghao Liu$^2$, Brian Jalaian$^3$, Jundong Li$^1$}
\affiliation{%
  \institution{$^1$University of Virginia \country{USA}} 
  \institution{$^2$University of Georgia \country{USA}} 
  \institution{$^3$U.S. Army Research Laboratory \country{USA}}}
\email{{yd6eb,jundong}@virginia.edu,ninghao.liu@uga.edu,brian.a.jalaian.civ@mail.mil}

\renewcommand{\shortauthors}{Yushun Dong, Ninghao Liu, Brian Jalaian, \& Jundong Li}

\begin{abstract}

Graph Neural Networks (GNNs) have shown superior performance in analyzing attributed networks in various web-based applications such as social recommendation and web search. Nevertheless, in high-stake decision-making scenarios such as online fraud detection, there is an increasing societal concern that GNNs could make discriminatory decisions towards certain demographic groups. 
Despite recent explorations on fair GNNs, these works are tailored for a specific GNN model. However, myriads of GNN variants have been proposed for different applications, and it is costly to fine-tune existing debiasing algorithms for each specific GNN architecture. 
Different from existing works that debias GNN models, we aim to debias the input attributed network to achieve fairer GNNs through feeding GNNs with less biased data. 
Specifically, we propose novel definitions and metrics to measure the bias in an attributed network, which leads to the optimization objective to mitigate bias. We then develop a framework EDITS to mitigate the bias in attributed networks while maintaining the performance of GNNs in downstream tasks. 
EDITS works in a model-agnostic manner, i.e., it is independent of any specific GNN. Experiments demonstrate the validity of the proposed bias metrics and the superiority of EDITS on both bias mitigation and utility maintenance. 
Open-source implementation: https://github.com/yushundong/EDITS.

\end{abstract}

%   <concept>
%       <concept_id>10010147.10010257</concept_id>
%       <concept_desc>Computing methodologies~Machine learning</concept_desc>
%       <concept_significance>500</concept_significance>
%       </concept>
%   <concept>
%       <concept_id>10010405.10010455</concept_id>
%       <concept_desc>Applied computing~Law, social and behavioral sciences</concept_desc>
%       <concept_significance>500</concept_significance>
%       </concept>
%  </ccs2012>

% \ccsdesc[500]{Computing methodologies~Machine learning}
% \ccsdesc[500]{Applied computing~Law, social and behavioral sciences}

\begin{CCSXML}
<ccs2012>
  <concept>
      <concept_id>10010147.10010257</concept_id>
      <concept_desc>Computing methodologies~Machine learning</concept_desc>
      <concept_significance>500</concept_significance>
      </concept>
 </ccs2012>
\end{CCSXML}

\ccsdesc[500]{Computing methodologies~Machine learning}


\keywords{graph neural networks, algorithmic fairness, data bias}

\maketitle
{\fontsize{8pt}{8pt} \selectfont  % 8pt
\textbf{ACM Reference Format:}\\
Yushun Dong, Ninghao Liu, Brian Jalaian, Jundong Li. 2022. EDITS: Modeling and Mitigating Data Bias for Graph Neural Networks. In \textit{Proceedings of ACM Web Conference 2022 (WWW ’22), April 25–29, 2022, Virtual Event, Lyon, France.} ACM, New York, NY, USA, 11 pages. https://doi.org/10.1145/3485447.\\3512173}

\section{Introduction}
\label{intro}

Attributed networks are ubiquitous in a plethora of web-related applications including online social networking~\cite{tang2009relational}, web advertising~\cite{yang2022wtagraph}, and news recommendation~\cite{qian2020interaction}.
To better understand these networks, various graph mining algorithms have been proposed. In particular, the recently emerged Graph Neural Networks (GNNs) have demonstrated superior capability of analyzing attributed networks in various tasks, such as node classification~\cite{kipf2016semi,velickovic2017graph} and link prediction~\cite{DBLP:conf/nips/ZhangC18,kipf2016variational}. Despite the superior performance of GNNs, they usually do not consider fairness issues in the learning process~\cite{dai2020say}.
Extensive research efforts have shown that many recently proposed GNNs~\cite{dai2020say,shumovskaia2021linking,xu2021towards} could make biased decisions towards certain demographic groups determined by sensitive attributes such as gender~\cite{ekstrand2021exploring} and political ideology~\cite{pariser2011filter}. 
For example, e-commerce platforms generate a huge amount of user activity data, and such data is often constructed as a large attributed network in which entities (e.g., buyers, sellers, and products) are nodes while activities between entities (e.g.., purchasing and reviewing) are edges. To prevent potential losses, fraud entities (e.g., manipulated reviews and fake buyers) need to be identified on these platforms, and GNNs have become the prevalent solution to achieve such goal~\cite{dou2020enhancing,liu2021pick}. Nevertheless, GNNs may have the risk of using sensitive information (e.g., race and gender) to identify fraud entities, yielding inevitable discrimination.
%
% in many e-commerce platforms, different entities (e.g., users and items) are connected as an attributed network based on activities between entities (e.g., purchasing). To prevent potential loss, fraud entities should be identified and banned~\cite{herbrich2017machine}. Nevertheless, GNNs may have the risk of using sensitive information (e.g., race and gender) to identify fraud entities, yielding inevitable discrimination.
% 
Therefore, it is a crucial problem to mitigate bias in these network-based applications.

Various efforts have been made to mitigate the bias exhibited in graph mining algorithms. For example, in online social networks, random walk algorithms can be modified via improving the appearance rate of minorities~\cite{rahman2019fairwalk, burke2017balanced}; adversarial learning is another popular approach, which aims to learn node embeddings that are not distinguishable on sensitive attributes~\cite{DBLP:conf/icml/BoseH19, masrour2020bursting}. Some recent efforts have also been made to mitigate bias in the outcome of GNNs. For example, adversarial learning can also be adapted to GNNs for outcome bias mitigation~\cite{dai2020say}. Nevertheless, existing approaches to debias GNN outcomes are tailored for a specific GNN model on a certain downstream task. In practical scenarios, different applications could adopt different GNN variants~\cite{kipf2016semi,hamilton2017inductive}, and it is costly to train and fine-tune the debiasing approaches based on diverse GNN backbones. As a consequence, to mitigate bias more efficiently for different GNNs and tasks, developing a one-size-fits-all approach becomes highly desired. Then the question is: how can we perform debiasing regardless of specific GNNs and downstream tasks? Considering that a model trained on biased datasets also tends to be biased~\cite{zemel2013learning,dai2020say,beutel2017data}, directly debiasing the dataset itself can be a straightforward solution. There are already debiasing approaches modifying original datasets via perturbing data distributions or reweighting the data points in the dataset~\cite{wang2019repairing,kamiran2012data,calders2009building}. These approaches obtain less biased datasets, which help to mitigate bias in learning algorithms. In this regard, considering that debiasing for different GNNs is costly, it is also highly desired to mitigate the bias in attributed networks before they are fed into GNNs. Nevertheless, to the best of our knowledge, despite its fundamental importance, no existing literature has taken such a step forward. 
% This motivates us to develop a principled debiasing framework for GNNs in the preprocessing stage of attributed networks in this paper.

% 在pre-processing的过程中提升fairness是十分困难的，因为这需要我们scrutinize产生fairness的原因。首先attributes自身是存在bias的；但是结构上也会有bais。比如谁谁就说过会这样。(这里我们给出一个toy example证明这件事情)

In this paper, we make an initial investigation on debiasing attributed networks towards more fair GNNs. Specifically, we tackle the following challenges.
(1) \emph{\textbf{Data Bias Modeling}}.
Traditionally, bias modeling is coupled with the outcome of a specific GNN~\cite{dai2020say}. Based on the GNN outcome, bias can be modeled via different fairness notions, e.g., \textit{Statistical Parity}~\cite{dwork2012fairness} and \textit{Equality of Opportunity}~\cite{DBLP:conf/nips/HardtPNS16}, to determine whether the outcome is discriminatory towards some specific demographic groups. Nevertheless, if debiasing is carried out directly based on the input attributed networks instead of the GNN outcome, the first and foremost challenge is how to appropriately model such data bias. 
(2) \emph{\textbf{Multi-Modality Debiasing}}. In fact, attributed networks contain both graph structure and node attribute information. Correspondingly, bias may exist with diverse formats across different data modalities. In this regard, how to debias attributed networks that have different data modalities is the second challenge that needs to be tackled.
%Such a data modality difference naturally makes attributed networks contain heterogeneous information sources.
(3) \emph{\textbf{Model-Agnostic Debiasing}}. 
Existing GNN debiasing approaches require the outcome of a specific GNN for objective function optimization during training. Different from these approaches, model-agnostic debiasing for GNNs should not rely on any specific GNN, as our goal is to develop a one-size-fits-all data debiasing approach to benefit various GNNs. Clearly, such model-agnostic debiasing could have better generalization capability but becomes much more difficult compared with the model-oriented GNN debiasing approaches. Nevertheless, the ultimate goal of debiasing is still to ensure the GNN outcome does not exhibit any discrimination. Such a contradiction poses the challenge of how to properly formulate a debiasing objective that can be universally applied to different GNNs in downstream tasks.
%is consistent with the goal of mitigating bias existing in the GNN output in downstream tasks. 

% is with higher generalization requirement and thus much more difficult compared with the model-oriented debiasing of the two baselines.

%因此，本文给出一种preprocessing的方法来做attributed network的debias。

To tackle the challenges above, we present novel data bias modeling approaches and a principled debiasing framework named EDITS (mod\underline{E}ling an\underline{D} m\underline{I}tigating da\underline{T}a bia\underline{S}) to achieve model-agnostic attributed network debiasing for GNNs. Specifically, we first carry out preliminary analysis to illustrate how bias exists in the two data modalities of an attributed network (i.e., node attributes and network structure) and affects each other in the information propagation of GNNs. Then, we formally define \textit{attribute bias} and \textit{structural bias}, together with the corresponding metrics for data bias modeling. Besides, we formulate the problem of debiasing attributed networks for GNNs, and propose a novel framework named EDITS for bias mitigation. It is worth mentioning that EDITS is model-agnostic for GNNs. In other words, our goal is to obtain less biased attributed networks for the input of any GNNs. 
%
% Meanwhile, theoretical analysis is conducted to build connections between the two kinds of bias, which reveals the essence of EDITS. 
%
Finally, empirical evaluations on both synthetic and real-world datasets corroborate the validity of the proposed bias metrics and the effectiveness of EDITS. Our contributions are summarized as:
\begin{itemize}[topsep=0pt]
    \item \textbf{Problem Formulation.} We formulate and make an initial investigation on a novel research problem: debiasing attributed networks for GNNs based on the analysis of the information propagation mechanism.
    
    \item \textbf{Metric and Algorithm Design.} We design novel bias metrics for attributed network data, and propose a model-agnostic debiasing framework named EDITS to mitigate the bias in attributed networks before they are fed into any GNNs.
    
    \item \textbf{Experimental Evaluation.} We conduct comprehensive experiments on both synthetic and real-world datasets to verify the validity of the proposed bias metrics and the effectiveness of the proposed framework EDITS.
\end{itemize}

\vspace{-1mm}
\section{Preliminary Analysis}
\label{investigation}
% handling the three challenges mentioned in Sec. \ref{intro}
We provide two exemplary cases to show how the two data modalities of an attributed network (i.e., node attribute and network structure) introduce bias in information propagation -- the most common operation in GNNs. These two cases also bring insights on tackling the three challenges mentioned in Sec.~\ref{intro}.
Specifically, two synthetic datasets are generated with either biased node attribute or network structure, and then attributes are propagated across the network structure to show how bias is introduced in GNNs.
Here we consider the attribute distribution difference between different demographic groups as the bias in attribute, while the group membership distribution difference of the neighbors for nodes between different demographic groups is regarded as the bias in network structure. Such bias in attribute and structure can be regarded as the bias that existed in two data modalities in an attributed network. It should be noted that using distribution difference to define the level of bias is consistent with many algorithmic fairness studies~\cite{dwork2012fairness,zemel2013learning},
Now we explain how the synthetic datasets are generated.
We assume the \textit{sensitive attribute} is gender, and 1,000 nodes are generated with half males (blue) and half females (orange) for both cases. In addition to the sensitive attribute, each node is with an extra two-dimensional attribute vector, which will be initialized and fed as input for information propagation. To introduce bias to either of the data modalities, different strategies are adopted to generate the attribute vector and the network structure.
To study how the two data modalities introduce bias in information propagation, we compare the distribution difference of attributes between groups before and after the propagation mechanism in GCN~\cite{kipf2016semi}.

% here we regard the distribution difference of attribute value and edge existence between groups as the bias in attribute and structure, respectively. In our two cases, we adopted either biased attribute or structure. Without loss of generality, we assume the \textit{sensitive attribute} is gender, and a synthetic network with 1,000 nodes is generated with half males (blue) and half females (orange) for both cases. In addition to the sensitive attribute, we also generate a two-dimensional attribute vector with continuous values following different strategies in the two cases for information propagation. Accordingly, we study the corresponding attribute distribution after information propagation.
% %
% Here the propagation mechanism of GCN~\cite{kipf2016semi} (i.e., multiply attribute matrix by the symmetrically normalized adjacency matrix) is applied in both cases to propagate the attribute information following the network structure.

\begin{figure}[!t]
    \vspace{-2mm}
    \centering
    \subfloat[Biased attributes]{
        \includegraphics[width=0.145\textwidth]{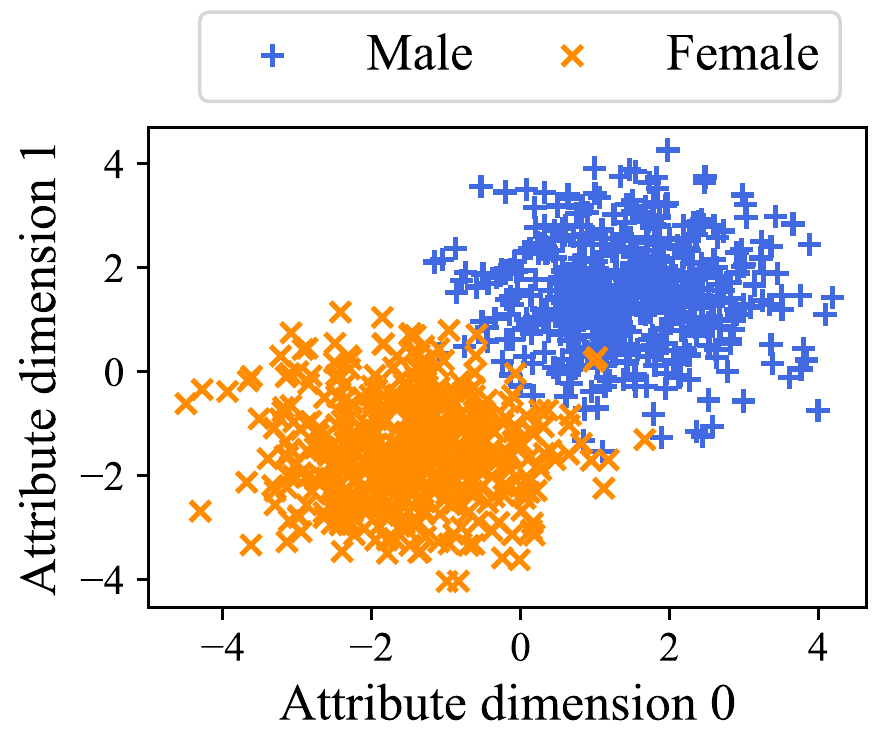}
        \label{dis_1_1}
    }
    \subfloat[Unbiased structure]{
        \includegraphics[width=0.145\textwidth]{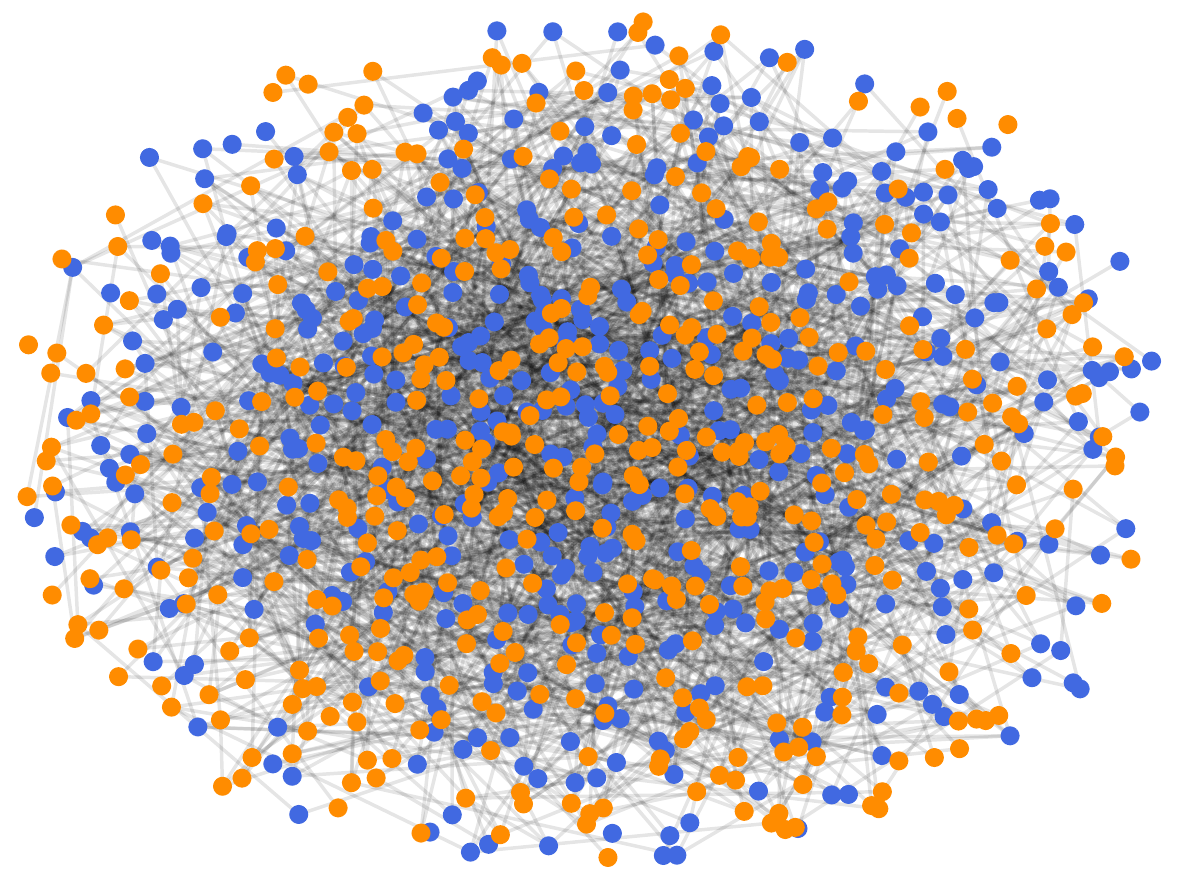}
        \label{st_1}
    }
    \subfloat[After propagation]{
        \includegraphics[width=0.145\textwidth]{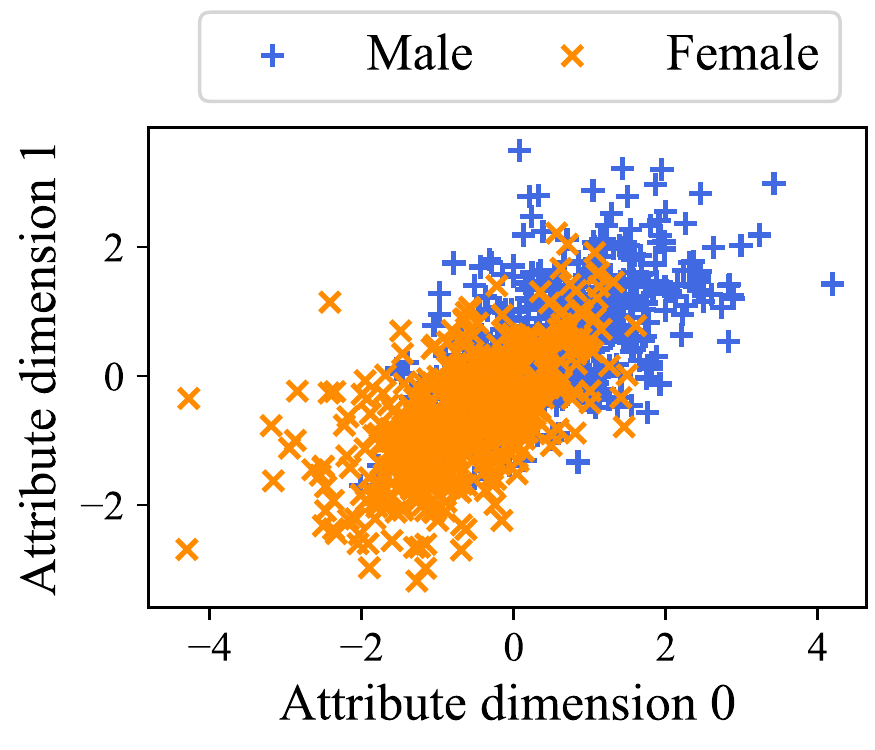}
        \label{dis_1_2}
    }
    \vspace{1.8pt}
    \subfloat[Unbiased attributes]{
        \includegraphics[width=0.145\textwidth]{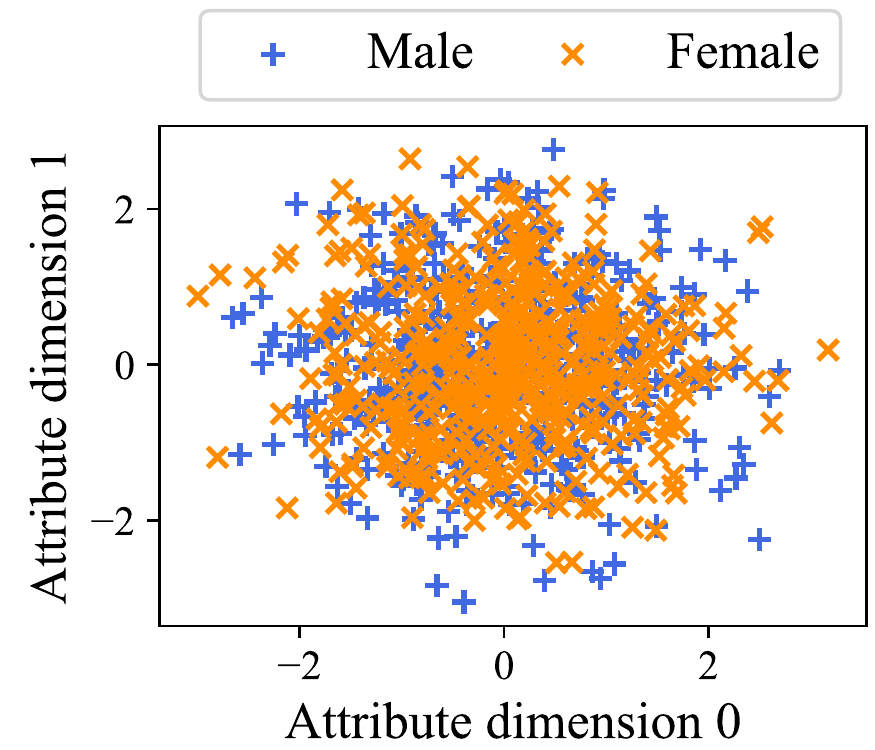}
        \label{dis_2_1}
    }
    \subfloat[Biased structure]{
        \includegraphics[width=0.145\textwidth]{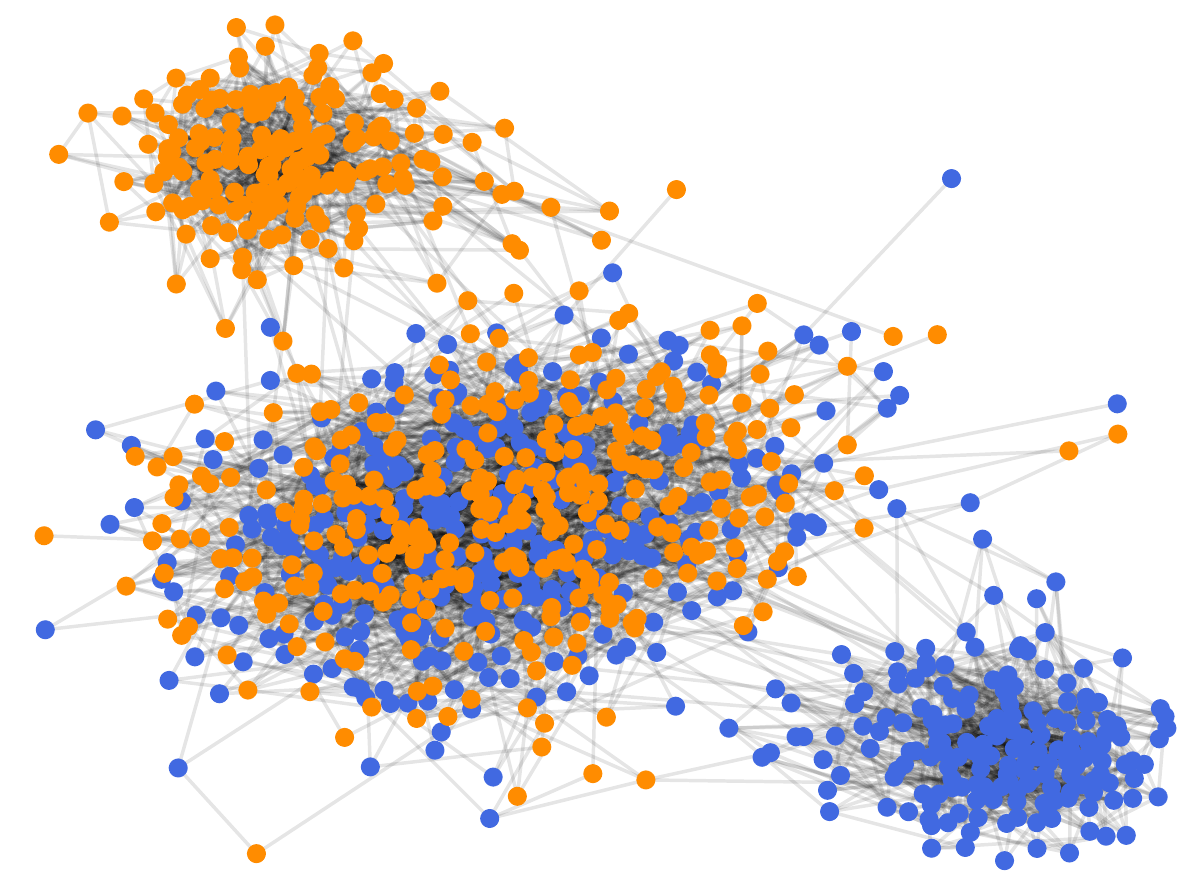}
        \label{st_2}
    }
    \subfloat[After propagation]{
        \includegraphics[width=0.145\textwidth]{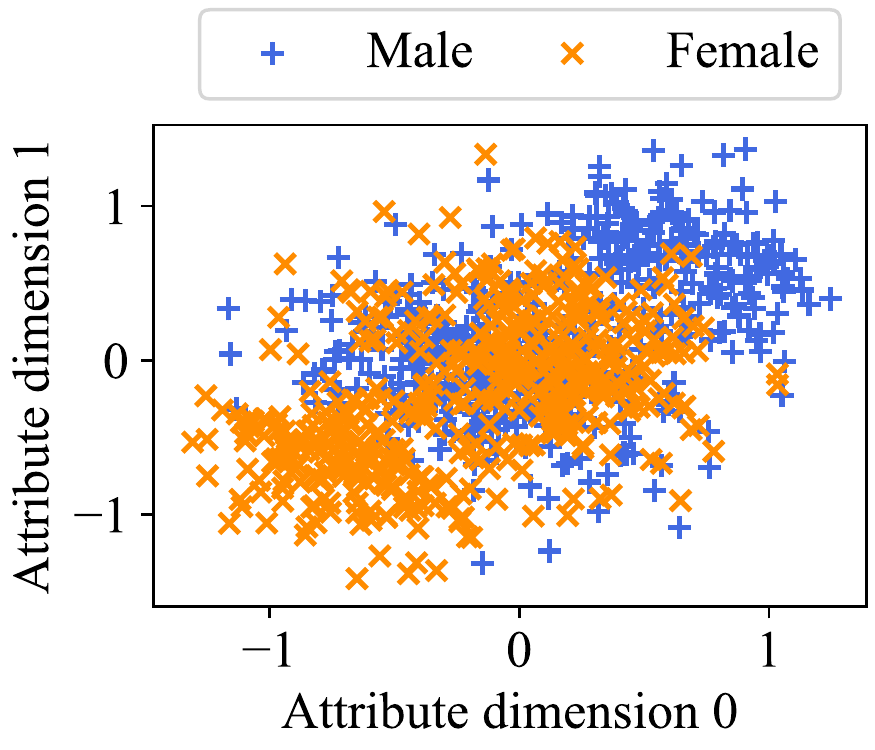}
        \label{dis_2_2}
    }
    \vspace{-4mm}
    \caption{Two exemplary cases illustrating how bias in the two data modalities of an attributed network introduce bias in GNN information propagation. Here (c) is the node attribute distribution after propagation with biased node attributes (a) and unbiased network structure (b); while (f) is the attribute distribution after propagation with unbiased node attributes (d) and biased network structure (e).}
        % \vspace{-6mm}
        \vspace{-6mm}
    \label{bias_incorporating}
\end{figure}

\noindent \textbf{Case 1: Biased attributes and unbiased structure.}
In this case, we generate biased two-dimensional attribute vectors for nodes from the two groups (i.e., males and females) and unbiased network structure. Specifically, biased attributes at each dimension is generated independently with Gaussian distribution $\mathcal{N}$(-1.5, $1^2$) for female and $\mathcal{N}$(1.5, $1^2$) for male. The distributions are shown in Fig.~(\ref{dis_1_1}). We then introduce how an unbiased network structure is generated. For each node in an unbiased network structure, the expected membership ratio of any group in its neighbor node set should be independent of the membership of the node itself. In this regard, we generate unbiased network structure via \textit{random graph} model with edge formation probability as $2 \times 10^{-3}$. The visualization of the network is presented in Fig.~(\ref{st_1}). The attribute distribution after information propagation according to the network structure is shown in Fig.~(\ref{dis_1_2}).
Comparing Fig.~(\ref{dis_1_1}) (attribute distribution before propagation) with (\ref{dis_1_2}) (attribute distribution after propagation), we can see the unbiased structure helps mitigate the original attribute bias after attributes are propagated according to the network structure. This not only implies that the attribute distribution difference between groups is a vital source of bias, but also demonstrates that unbiased structure helps mitigate bias in attributes after the information propagation process.

\noindent \textbf{Case 2: Unbiased attributes and biased structure.}
In this case, unbiased attributes are generated independently at each dimension with $\mathcal{N}$(0, $1^2$) for both males and females. The distributions are shown in Fig.~(\ref{dis_2_1}). The biased network structure is generated as follows. For each node, we sum up its attribute values. Then, we rank all nodes in descending order according to the summation of attribute values. After that, given a threshold integer $t$, for the top-ranked $t$ males and bottom-ranked $t$ females, we assume that they form two separated communities. The two communities are shown as the bottom right community (males) and the upper left community (females) in Fig.~(\ref{st_2}). We generate edges via \textit{random graph} model with edge formation probability as $5 \times 10^{-2}$ within each community. Similarly, the rest nodes form the third community via \textit{random graph} model with edge formation probability as $1 \times 10^{-2}$. We also generate edges between nodes from the male (or female) community and the third community with the probability of $2 \times 10^{-4}$. In this way, we introduce bias in network structure. The final network is presented in Fig.~(\ref{st_2}). The attribute distribution after propagation according to the network structure is shown in Fig.~(\ref{dis_2_2}).
Comparing Fig.~(\ref{dis_2_1}) with (\ref{dis_2_2}), we find that even if the original attributes are unbiased, the biased structure still turns the attributes into biased ones after information propagation. This implies that the bias contained in the network structure is also a significant source of bias.
%(attribute distribution before propagation) 
%(attribute distribution after propagation)

Based on the discussions, we draw three preliminary conclusions to help us tackle the challenges in Sec.~\ref{intro}.
(1) For \emph{\textbf{Data Bias Modeling}}, bias in attributes can be modeled based on the difference of attribute distribution between two groups. Also, bias in network structure can be modeled based on the difference of attribute distribution between two groups after information propagation.
(2) For \emph{\textbf{Multi-Modality Debiasing}} in an attributed network, at least two debiasing processes should be carried out targeting the two data modalities (i.e., attributes and structure).
(3) For \emph{\textbf{Model-Agnostic Debiasing}}, if the attribute distributions between groups can be less biased both before and after information propagation, the learned node representations tend to be indistinguishable between groups. Then GNNs trained on such data could also be less biased.
%Such difference mitigation between distributions from two groups is broadly consistent with bias mitigation~\cite{dwork2012fairness}.}
% since information propagation is also the core operation in GNNs

% \vspace{-2mm}
\section{Modeling Data Bias for GNNs}
\label{metrics}
In this section, we define \textit{attribute bias} and \textit{structural bias} in attributed networks together with their metrics. 
For the sake of simplicity, we focus on binary sensitive attribute and leave the generalization to non-binary cases in the Appendix.
% Only binary sensitive attributes are considered here, and generalization will be discussed in Sec.~\ref{metrics}.
%
% Without loss of generality, only binary sensitive attributes are considered, which can be easily generalized to more complicated scenarios. 
%Theoretical analysis corresponding to the metrics will be presented in Section~\ref{theo}.
%
Theoretical analysis of our proposed metrics is also presented in the Appendix.

% \vspace{-2mm}
\subsection{Preliminaries}
In this paper, without further specification, bold uppercase letters (e.g., $\mathbf{X}$), bold lowercase letters (e.g., $\mathbf{x}$), and normal lowercase letters (e.g., $x$) represent matrices, vectors, and scalars, respectively. For any matrix, e.g., $\mathbf{X}$, we use $\mathbf{X}_i$ denote its $i$-th row.
% , $\mathbf{X}_{ij}$ to   and its ($i$,$j$)-th entry, respectively

Let $\mathcal{G}$ = ($\mathbf{A}$, $\mathbf{X}$) be an undirected attributed network. Here $\mathbf{A} \in \mathbb{R}^{N \times N}$ is the adjacency matrix, and $\mathbf{X} \in \mathbb{R}^{N \times M}$ is the node attribute matrix, where $N$ is the number of nodes and $M$ is the attribute dimension. Let a diagonal matrix $\mathbf{D}$ be the degree matrix of $\mathbf{A}$, where its ($i$,$i$)-th entry $\mathbf{D}_{i,i} = \sum_{j} \mathbf{A}_{i,j}$, and $\mathbf{D}_{i,j} = 0$ ($i\neq j$). $\mathbf{L} = \mathbf{D} - \mathbf{A}$ is the graph Laplacian matrix. Denote the normalized adjacency matrix and the normalized Laplacian matrix as $\mathbf{A}_{\text{norm}} = \mathbf{D}^{- \frac{1}{2}} \mathbf{A} \mathbf{D}^{- \frac{1}{2}}$ and $\mathbf{L}_{\text{norm}} = \mathbf{D}^{- \frac{1}{2}} \mathbf{L} \mathbf{D}^{- \frac{1}{2}}$. $|.|$ is the absolute value operator.

\subsection{Definitions of Bias}
\label{definitions}
We consider two types of bias on attributed networks, i.e., attribute bias and structural bias. We first define attribute bias as follows.

\begin{myDef}\label{defn:attribute_bias}
\textbf{Attribute bias.} Given an undirected attributed network $\mathcal{G}$ = ($\mathbf{A}$, $\mathbf{X}$) and the group indicator (w.r.t. the sensitive attribute) for each node $\mathbf{s} = [s_1, s_2, ..., s_N]$, where $s_i \in \{0, 1\}$ ($1 \leq i \leq N$). For any attribute, if its value distributions between different demographic groups are different, then attribute bias exists in $\mathcal{G}$.
\end{myDef}

Besides, as shown in the second example in Sec.~\ref{investigation}, bias can also emerge after attributes are propagated in the network even when original attributes are unbiased. Therefore, an intuitive idea to identify structural bias is to check whether information propagation in the network introduces or exacerbates bias~\cite{jalali2020information}. Formally, we define structural bias on attributed networks as follows.

\begin{myDef}\label{defn:structural_bias}
\textbf{Structural bias.} Given an undirected attributed network $\mathcal{G}$ = ($\mathbf{A}$, $\mathbf{X}$) and the corresponding group indicator (w.r.t. sensitive attribute) for each node $\mathbf{s} = [s_1, s_2, ..., s_N]$, where $s_i \in \{0, 1\}$ ($1 \leq i \leq N$). For the attribute values propagated w.r.t. $\mathbf{A}$, if their distributions between different demographic groups are different at any attribute dimension, then structural bias exists in $\mathcal{G}$.
\end{myDef}

%Definition~\ref{defn:attribute_bias} and Definition~\ref{defn:structural_bias} aim to distinguish whether attribute bias and structural bias exists in a certain network.
Apart from these definitions, it is also necessary to quantitatively measure the attribute bias and structural bias. In the sequel, we introduce our proposed metrics for the two types of bias.

\subsection{Bias Metrics}
\label{metrics}
% 实在不行就只能换成normalize过的A，场景限制在无向图，然后用L来证明了
% Now we introduce the metrics to quantitatively measure attribute bias and structural bias for an undirected attributed network $\mathcal{G}$. 

Here we take the first step to define metrics for both \textit{attribute bias} and \textit{structural bias} for an undirected attributed network $\mathcal{G}$.
%
% We defer a detailed theoretical analysis that builds the connections between these two bias metrics in Section~\ref{theo}.

\noindent \textbf{Attribute bias metric.} Let $\mathbf{X}_{\text{norm}} \in \mathbb{R}^{N \times M}$ be the normalized attribute matrix. For the $m$-th dimension ($1 \leq m \leq M$) of $\mathbf{X}_{\text{norm}}$, we use $\mathcal{X}^{0}_m$ and $\mathcal{X}^{1}_m$ to denote attribute value set for nodes with $s_{i}=0$ and $s_{i}=1$ ($1 \leq i \leq N$). Then, attributes of all nodes can be divided into tuples: $\mathcal{X}_{total} = \{ (\mathcal{X}^{0}_1, \mathcal{X}^{1}_1), (\mathcal{X}^{0}_2, \mathcal{X}^{1}_2), ..., (\mathcal{X}^{0}_M, \mathcal{X}^{1}_M)\}$. We measure attribute bias with Wasserstein-1 distance~\cite{villani2021topics} between the distributions of the two groups:
\begin{align}
\label{bias_1_formulation}
b_{\text{attr}} = \frac{1}{M} \sum_{m} W (pdf(\mathcal{X}^{0}_m), pdf(\mathcal{X}^{1}_m)).
\end{align}
Here $pdf (\cdot)$ is the probability density function for a set of values, and $W$(., .) is the Wasserstein distance between two distributions. Intuitively, $b_{\text{attr}}$ describes the average Wasserstein-1 distance between attribute distributions of different groups across all dimensions. It should be noted that taking the distribution difference between demographic groups as the indication of bias is in align with many existing algorithmic fairness studies~\cite{zemel2013learning,DBLP:conf/icml/BoseH19,dai2020say}.

\begin{figure*}[!t]
\vspace{-3mm}
    \centering
    \includegraphics[width=0.65\textwidth]{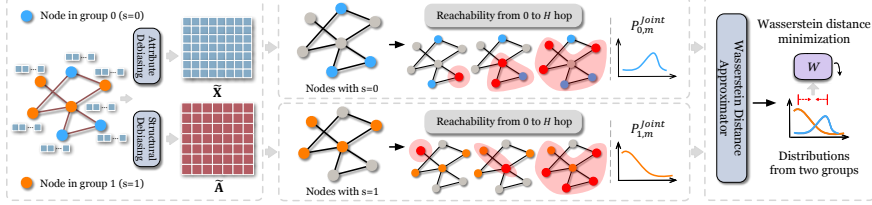}
    \vspace{-3mm}
    % \caption{An illustration of the proposed framework EDITS with \textbf{$H=2$}: Wasserstein Distance Approximator yields the approximated Wasserstein distance between $P_{0,m}^{Joint}$ and $P_{1,m}^{Joint}$; Attribute Debiasing and Structural Debiasing are optimized towards less biased $\mathbf{\tilde{X}}$ and $\mathbf{\tilde{A}}$ based on the Wasserstein distance approximation.}
    \caption{An illustration of EDITS with \textbf{$H=2$}: Wasserstein Distance Approximator yields the approximated Wasserstein distance between $P_{0,m}^{Joint}$ and $P_{1,m}^{Joint}$; Attribute Debiasing and Structural Debiasing are optimized towards less biased $\mathbf{\tilde{X}}$ and $\mathbf{\tilde{A}}$.}
    \vspace{-4mm}
    \label{framework}
\end{figure*}

\noindent \textbf{Structural bias metric.} 
% Define $\mathbf{P} = \alpha \mathbf{A} + (1- \alpha) \mathbf{D}$ and the corresponding normalized $\mathbf{P}_{\text{norm}} = \mathbf{D}^{- \frac{1}{2}}  \mathbf{P}  \mathbf{D}^{- \frac{1}{2}} = \alpha \textbf{A}_{\text{norm}} + (1 - \alpha) \textbf{I}$. 
As illustrated in Sec.~\ref{investigation}, the key mechanism of GNNs is information propagation, during which the structural bias could be introduced.
% Here we inherit the similar idea to define structural bias.
%
Let $\mathbf{P}_{\text{norm}} = \alpha \textbf{A}_{\text{norm}} + (1 - \alpha) \textbf{I}$. 
Here $\mathbf{P}_{\text{norm}}$ can be regarded as a normalized adjacency matrix with re-weighted self-loops, where $\alpha \in [0,1]$ is a hyper-parameter. Before measuring structural bias, we define the \textit{propagation matrix} $\mathbf{M}_H \in \mathbb{R}^{N \times N}$ as:
\begin{align}\label{eq:Mh}
\mathbf{M}_{H} = \beta_1 \mathbf{P}_{\text{norm}} + \beta_2 \mathbf{P}_{\text{norm}}^2 + ... + \beta_H \mathbf{P}_{\text{norm}}^H,
\end{align}
where $\beta_h$ ($1 \leq h \leq H$) is re-weighting parameters. The rationale behind the formulation above is to measure the aggregated reaching likelihood from each node to other nodes within a distance of $H$. To achieve localized effect for each node, a desired choice is to let $\beta_1 \geq \beta_2 \geq ... \geq \beta_H$, i.e., emphasizing short-distance terms and reducing the weights of long-distance terms. For example, assume $H=3$, then the value $(\mathbf{M}_{3})_{i,j}$ is the aggregated reaching likelihood from node $i$ to node $j$ within 3 hops with re-weighting parameters being $\beta_1$, $\beta_2$ and $\beta_3$.
%for the term of 1-hop, 2-hop and 3-hop, respectively. 
%
Also, given attributes $\mathbf{X}_{\text{norm}}$, we define the \textit{reachability matrix} $\mathbf{R} \in \mathbb{R}^{N \times M}$ as $\mathbf{R} = \mathbf{M}_{H} \mathbf{X}_{\text{norm}}$. Intuitively, $\mathbf{R}_{i, m}$ is the aggregated reachable attribute value for attribute $m$ of node $i$. 
We utilize $\mathcal{R}^{0}_m$ and $\mathcal{R}^{1}_m$ to represent the set of values of the $m$-th dimension in $\mathbf{R}$ for nodes with $s_{i}=0$ and $s_{i}=1$ ($1 \leq i \leq N$). The entries in $\mathbf{R}$ can also be divided into tuples according to attribute dimensions: $\mathcal{R}_{total} = \{ (\mathcal{R}^{0}_1, \mathcal{R}^{1}_1), (\mathcal{R}^{0}_2, \mathcal{R}^{1}_2), ..., (\mathcal{R}^{0}_M, \mathcal{R}^{1}_M)\}$. We define structural bias as:
\begin{align}
\label{bias_2_formulation}
b_{\text{stru}} = \frac{1}{M} \sum_{m} W (pdf(\mathcal{R}^{0}_m), pdf(\mathcal{R}^{1}_m)).
\end{align}
Here $b_{\text{stru}}$ is defined in a similar way as $b_{\text{attr}}$, except that the former uses $\mathcal{R}^{0}_m$ and $\mathcal{R}^{1}_m$ instead of $\mathcal{X}^{0}_m$ and $\mathcal{X}^{1}_m$. In this way, structural bias $b_{\text{stru}}$ describes the average difference between aggregated attribute distributions of different groups after several rounds of propagation.

% \noindent\textbf{Extension to Non-Binary Sensitive Attributes.}
% %
% Although we assume the sensitive attribute is binary to simplify the discussion, the proposed metrics can be easily generalized to any non-binary sensitive attributes. In fact, such extension is quite straightforward. For non-binary categorical sensitive attributes, we leverage the average of $b_{\text{attr}}$ between each pair of demographic groups to measure the attribute bias, and structural bias extension can follow a similar way. For continuous sensitive attributes, we divide the value range of sensitive attributes into multiple bins. The number of each bin can be the class of the nodes that fall into the bin. Then the extension for non-binary categorical sensitive attributes can be directly applied.

% each Wasserstein distance term in Eq. (\ref{bias_1_formulation}) and Eq. (\ref{bias_2_formulation}) can be replaced with ; for continuous sensitive attributes, ;
% each Wasserstein distance term in Eq. (\ref{bias_1_formulation}) and Eq. (\ref{bias_2_formulation}) can be replaced with the total distance between all group pairs, where the framework for optimization in Sec. \ref{framework_intro} is also applicable. Thus, in the rest of the paper, we still focus on binary sensitive attributes for illustration.

\subsection{Problem Statement}
% 首先强调我们提出的metric和下有任务公平性之间的联系？？？

% 围绕如何降低value of proposed bias metrics

% 但最后的RQ之一需要验证我们的metric是不是跟传统的metric在downstream tasks上consistent

Based on the definitions and metrics in Sec.~\ref{definitions} and \ref{metrics}, we argue that if both $b_{attr}$ and $b_{stru}$ are reduced, bias in an attributed network can be mitigated. As a result, if GNNs are trained on such data, the bias issues in downstream tasks could also be alleviated. Formally, we define the debiasing problem as follows.

\begin{myDef1}
\label{problem}
\textbf{Debiasing attributed networks for GNNs.} Given an attributed network $\mathcal{G}$ = ($\mathbf{A}$, $\mathbf{X}$), our goal is to debias $\mathcal{G}$ by reducing $b_{attr}$ and $b_{stru}$ to obtain $\mathcal{\tilde{G}}$ = ($\mathbf{\tilde{A}}$, $\mathbf{\tilde{X}}$), so that the bias of GNNs trained on $\mathcal{\tilde{G}}$ is mitigated. The debiasing is independent of any specific GNNs.

% does not consider the structure or parameters of GNNs.
\end{myDef1}

%in a model-agnostic way 

%\textbf{Debiasing attributed networks for GNNs.} Given an attributed network $\mathcal{G}$ = ($\mathbf{A}$, $\mathbf{X}$), our goal is to debias $\mathcal{G}$ according to Definition \ref{defn:attribute_bias} and \ref{defn:structural_bias} (i.e., to mitigate $b_{attr}$ and $b_{stru}$) in a model-agnostic way to achieve $\mathcal{G}'$ = ($\mathbf{A}'$, $\mathbf{X}'$), in order to mitigate the bias of GNNs trained on $\mathcal{G}'$ in downstream tasks.

% In the following section, we discuss how we handle Problem~\ref{problem} with our proposed framework EDITS. We will also verify the consistency between our proposed bias metrics and traditional bias metrics (e.g., \textit{Statistical Parity} and \textit{Equality of Opportunity}) corresponding to specific downstream tasks in our experiments.

\section{Mitigating Data Bias for GNNs}
\label{framework_intro}
In this section, we discuss how to tackle Problem~\ref{problem} with our proposed framework EDITS. We focus on the binary sensitive attribute for the sake of simplicity and discuss the extension later. We first present an overview of EDITS, followed by the formulation of the objective function. Finally, we present the optimization process.

\subsection{Framework Overview}
An overview of the proposed framework EDITS is shown in Fig.~(\ref{framework}). Specifically, EDITS consists of three modules. The parameters of these three modules are optimized alternatively during training.
\begin{itemize}
    \item \textbf{\emph{Attribute Debiasing.}} This module learns a debiasing function $g_{\bm{\theta}}$ with learnable parameter $\bm{\theta} \in \mathbb{R}^{M}$. The debiased version of $\mathbf{X}$ is obtained as output where $\mathbf{\tilde{X}} = g_{\bm{\theta}} (\mathbf{X})$. 
    % In this paper, we implement $g_{\bm{\theta}}$ as a linear mapping function.
    
    \item \textbf{\emph{Structural Debiasing.}} This module outputs $\mathbf{\tilde{A}}$ as the debiased $\mathbf{A}$. Specifically, $\mathbf{\tilde{A}}$ is initialized with $\mathbf{A}$ at the beginning of the optimization process. The entries in $\mathbf{\tilde{A}}$ are optimized via gradient descent with clipping and binarization.
    
    \item \textbf{\emph{Wasserstein Distance Approximator.}} This module learns a function $f$ for each attribute dimension. Here $f$ is utilized to estimate the Wasserstein distance between the distributions of different groups for any attribute dimension.
    % of the attribute matrix and reachability matrix
\end{itemize}

\vspace{-0.1in}
\subsection{Objective Function}
% 介绍一下metric和方法的联系
In this subsection, we introduce the details of our framework. Following the Definition~\ref{defn:attribute_bias} and Definition~\ref{defn:structural_bias}, our goal is to reduce $b_{\text{attr}}$ and $b_{\text{stru}}$ simultaneously. For the ease of understanding, we first only consider the $m$-th attribute dimension as an example, and then extend it to all $M$ dimensions to obtain our final objective function.

Let $P_{0,m}$ and $P_{1,m}$ be the value distribution at the $m$-th attribute dimension in $\mathbf{X}$ for nodes with sensitive attribute $s=0$ and $s=1$, respectively. Denote $x_{0,m} \sim P_{0,m}^{(h)}$ and $x_{1,m} \sim P_{1,m}^{(h)}$ as two random variables drawn from the two distributions. Assume that we have a function $g_{\theta_m}: \mathbb{R} \rightarrow \mathbb{R}$ to mitigate attribute bias, where $1 \leq m \leq M$. For the $m$-th dimension, we denote $x_{0,m}^{(0)} = g_{\theta_m}(x_{0,m}) \sim P_{0,m}^{(0)}$ and $x_{1,m}^{(0)} = g_{\theta_m}(x_{1,m}) \sim P_{1,m}^{(0)}$ as the debiasing results for $x_{0,m}$ and $x_{1,m}$, respectively. Here the superscript $(0)$ indicates that no information propagation is performed in the debaising process. Correspondingly, when such operation is extended to all $M$ dimensions, we will have the debiased attribute matrix $\mathbf{\tilde{X}}$. Apart from the goal of mitigating attribute bias, we also want to mitigate structural bias. Let $\mathbf{\tilde{A}}$ be the adjacency matrix from the debiased network structure, and $\mathbf{\tilde{P}}_{\text{norm}}$ denotes the normalized $\mathbf{\tilde{A}}$ with re-weighted self-loops.
Information propagation with $h$ hops using the debiased adjacency matrix could be expressed as $\mathbf{\tilde{P}}_{\text{norm}}^{h} \mathbf{\tilde{X}}$, where $1 \leq h \leq H$.
Let $P_{0,m}^{(h)}$ and $P_{1,m}^{(h)}$ be the value distribution at the $m$-th column of $\mathbf{\tilde{P}}_{\text{norm}}^{h} \mathbf{\tilde{X}}$ for nodes with sensitive attribute $s=0$ and $s=1$, respectively. Denote $x_{0,m}^{(h)} \sim P_{0,m}^{(h)}$ and $x_{1,m}^{(h)} \sim P_{1,m}^{(h)}$ as two random variables drawn from the two distributions.
%
% Let $x_{0,m}^{(h)} \sim P_{0,m}^{(h)}$ and $x_{1,m}^{(h)} \sim P_{1,m}^{(h)}$ be the random variables of the value at the $m$-th column of $\mathbf{\tilde{P}}_{\text{norm}}^{h} \mathbf{\tilde{X}}$ for nodes with sensitive attribute $s=0$ and $s=1$, respectively.
%
We hope that $\mathbf{\tilde{A}}$ could mitigate structural bias. We combine attribute and structural debiasing as below.
%in the matrix $\mathbf{X}$ 
%, so here $x_{0,m}$ and $x_{1,m}$ are two random variables

Based on the random variables $x_{0,m}^{(0)}$ to $x_{0,m}^{(H)}$ and $x_{1,m}^{(0)}$ to $x_{1,m}^{(H)}$, we have $(H+1)$-dimensional vectors $\mathbf{x}_{0,m} = [x_{0,m}^{(0)}, x_{0,m}^{(1)}, ..., x_{0,m}^{(H)}]$ and $\mathbf{x}_{1,m} = [x_{1,m}^{(0)}, x_{1,m}^{(1)}, ..., x_{1,m}^{(H)}]$ following the joint distribution $P_{0,m}^{Joint}$ and $P_{1,m}^{Joint}$, respectively. To reduce both $b_{\text{attr}}$ and $b_{\text{stru}}$ at the $m$-th dimension, our goal is to minimize the Wasserstein distance between $P_{0,m}^{Joint}$ and $P_{1,m}^{Joint}$, which is formulated as $\min_{\theta_m, \mathbf{\tilde{A}}} W(P_{0,m}^{Joint}, P_{1,m}^{Joint})$.
% \begin{align}
% \label{wd_1}
%     \min_{\theta_m, \mathbf{\tilde{A}}} W(P_{0,m}^{Joint}, P_{1,m}^{Joint}),
% \end{align}
$W(P_{0,m}^{Joint}, P_{1,m}^{Joint})$ can be expressed as
\begin{align}
\label{wd_2}
    W(P_{0,m}^{Joint}, &P_{1,m}^{Joint}) =  \\  \notag
    &\inf_{\gamma \in \Pi(P_{0,m}^{Joint}, P_{1,m}^{Joint})} \mathbb{E}_{(\mathbf{x}_{0,m}, \mathbf{x}_{1,m}) \sim \gamma}[\|\mathbf{x}_{0,m}-\mathbf{x}_{1,m}\|_{1}].
\end{align}
Here $\Pi(P_{0,m}^{Joint}, P_{1,m}^{Joint})$ represents the set of all joint distributions $\gamma (\mathbf{x}_{0,m},\mathbf{x}_{1,m})$ whose marginals are $P_{0,m}^{Joint}$ and $P_{1,m}^{Joint}$, respectively. After considering all the $M$ dimensions, the overall objective is 
\begin{align}
\label{wd_3}
    \min_{\bm{\theta}, \mathbf{\tilde{A}}} \frac{1}{M} \sum_{1 \leq m \leq M} W(P_{0,m}^{Joint}, P_{1,m}^{Joint}).
\end{align}
It is non-trivial to optimize Eq. (\ref{wd_3}) as the infimum is intractable. Therefore, in the next subsection, we show how to convert it into a tractable optimization problem through approximation, which enables end-to-end gradient-based optimization.
%Here $\bm{\theta}$ is a vector consisting of all $\theta_m$ ($1 \leq m \leq M$).

% Although we assume the sensitive attribute is binary to simplify the discussion, but the introduced metrics can be easily generalized to any non-binary sensitive attribute. In fact, such extension is quite straightforward: each Wasserstein distance term in Eq. (\ref{bias_1_formulation}) and Eq. (\ref{bias_2_formulation}) can be replaced with the total distance between all group pairs, where the framework for optimization in Sec. \ref{framework_intro} is also applicable. Thus, in the rest of the paper, we still focus on binary sensitive attributes for illustration.

\subsection{Framework Optimization}
\label{optimization}
In this subsection, we introduce our optimization algorithm. For simplicity, first we still use the $m$-th attribute dimension in $\mathbf{X}$ to illustrate the idea. Considering the infimum in Wasserstein distance computation is intractable, we apply the Kantorovich-Rubinstein duality~\cite{villani2008optimal} to convert the problem of Eq. (\ref{wd_2}) as:
% In this subsection, we introduce how we optimize our objective function in Eq. (\ref{wd_3}). Considering the infimum included in $W(.,.)$ is intractable, the basic logic here is that we first approximate the Wasserstein distance in Eq. (\ref{wd_3}), then we minimize such approximation. We still use the $m$-th attribute dimension of $\mathbf{X}$ to introduce the idea. Specifically, the Wasserstein distance computation between $P_{0,m}^{Joint}$ and $P_{1,m}^{Joint}$ is given as Eq. (\ref{wd_2}). We utilize the Kantorovich-Rubinstein duality~\cite{villani2008optimal} to convert this problem into 
\begin{align}
    \label{wd_4}
    W(&P_{0,m}^{Joint}, P_{1,m}^{Joint}) = \\   \notag
    &\sup_{\|f\|_{L} \leq 1} \mathbb{E}_{\mathbf{x}_{0,m} \sim P_{0,m}^{Joint}}[f(\mathbf{x}_{0,m})] - \mathbb{E}_{\mathbf{x}_{1,m} \sim P_{1,m}^{Joint}}[f(\mathbf{x}_{1,m})].
\end{align}
Here $\|f\|_{L} \leq 1$ denotes that the supremum is taken over all 1-Lipschitz functions $f$ : $\mathbb{R}^{H+1} \rightarrow \mathbb{R}$.
%There is a solution of $f$ which maximizes Eq. (\ref{wd_4}) as $\mathbb{R}^{H+1}$ is a compact set~\cite{DBLP:journals/corr/ArjovskyCB17}. 
The problem can be solved by learning a neural network as $f$. Nevertheless, it is worth noting that the 1-Lipschitz function is difficult to obtain during optimization. 
%Therefore, here we extend the constraint from $||f||_{L} \leq 1$ to $||f||_{L} \leq k$ (here $k$ is a constant). 
Therefore, here we relax $\|f\|_{L} \leq 1$ to $\|f\|_{L} \leq k$ ($k$ is a constant). 
In this case, the left side of Eq. (\ref{wd_4}) also changes to $k W(P_{0,m}^{Joint}, P_{1,m}^{Joint})$. Then, the Wasserstein distance between $P_{0,m}^{Joint}$ and $P_{1,m}^{Joint}$ up to a multiplicative constant can be attained via:
\begin{align}
    \label{wd_5}
    &\max_{f_m \in \mathcal{F}} \mathbb{E}_{\mathbf{x}_{0,m} \sim P_{0,m}^{Joint}}[f_{m}(\mathbf{x}_{0,m})] - \mathbb{E}_{\mathbf{x}_{1,m} \sim P_{1,m}^{Joint}}[f_{m}(\mathbf{x}_{1,m})],
\end{align}
where $\mathcal{F}$ denotes the set of all $k$-Lipschitz functions (i.e., $\|f_{m}\|_{L} \leq k$, $f_m \in \mathcal{F}$). Then, extending Eq. (\ref{wd_5}) to all $M$ dimensions leads to our final objective function as:
\begin{align}
    \label{wd_6}
     \mathscr{L}_1 = \sum_{1 \leq m \leq M} \{ \mathbb{E}_{\mathbf{x}_{0,m} \sim P_{0,m}^{Joint}}[f_{m}(\mathbf{x}_{0,m})] - \mathbb{E}_{\mathbf{x}_{1,m} \sim P_{1,m}^{Joint}}[f_{m}(\mathbf{x}_{1,m})] \},
\end{align}
where $\{f_m: 1\leq m \leq M\} \subset \mathcal{F}$.
% instantiation
To model the function $f$ in Eq. (\ref{wd_6}), a single-layered neural network serves as the \textit{Wasserstein Distance Approximators} in Fig.~(\ref{framework}) to approximate each $f_m$ ($1 \leq m \leq M$), where the objective can be formulated as:
\begin{align}
    \label{wd_7}
     \max_{\{f_m: 1 \leq m \leq M\} \subset \mathcal{F}}  \mathscr{L}_1 \,\,.
\end{align}
The weights of neural networks are clipped within $[-c, c]$ ($c$ is a pre-defined constant), which has been proved to be a simple but effective way to enforce the Lipschitz constraint for every $f_{m}$~\cite{DBLP:journals/corr/ArjovskyCB17}. 
%
%For the debiasing function $g_{\theta}$ for attributes, we choose a linear function (\textit{Attribute debiasing}) here, i.e., $g_{\theta_m}(x_{s,m}) = \theta_m x_{s,m}$ ($s \in \{0, 1\}$). In matrix form, assume $\bm{\Theta}$ is a diagonal matrix with the $m$-th diagonal entry being $\theta_m$. The corresponding optimization goal for \textit{Attribute debiasing} is
For the \textit{Attribute Debiasing} module in Fig.~(\ref{framework}), we choose a linear function, i.e., $g_{\theta_m}(x_{s,m}) = \theta_m x_{s,m}$ ($s \in \{0, 1\}$). One advantage is that it acts as the role of feature re-weighting by assigning a feature weight for each attribute, which enables better interpretability for the debiased result. In matrix form, assume $\bm{\Theta}$ is a diagonal matrix with the $m$-th diagonal entry being $\theta_m$, we have $\mathbf{\tilde{X}} =g_{\bm{\theta}} (\mathbf{X})= \mathbf{X} \bm{\Theta}$. Then the optimization goal for \textit{attribute debiasing} is:
\begin{align}
    \label{wd_8}
     \min_{\bm{\Theta}}\, \mathscr{L}_1 + \mu_1 \|\mathbf{\tilde{X}} - \mathbf{X}\|_{F}^{2} + \mu_2 \|\bm{\Theta}\|_{1},
\end{align}
where $\mu_1$ and $\mu_2$ are hyper-parameters. The second term ensures that the debiased attributes after feature re-weighting are close to the original ones (i.e., preserve as much information as possible). The third term controls the sparsity of re-weighting parameters. 
For the \textit{Structural Debiasing} module in Fig.~(\ref{framework}), $\mathbf{\tilde{A}}$ is optimized through:
\begin{align}
    \label{wd_9}
     \min_{\mathbf{\tilde{A}}}\, \mathscr{L}_1 + \mu_3 \|\mathbf{\tilde{A}} - \mathbf{A}\|_{F}^{2} + \mu_4 \|\mathbf{\tilde{A}} \|_{1} \;\;\; s.t., \mathbf{\tilde{A}} = \mathbf{\tilde{A}}^{\top}.
\end{align}
%We clip the range of $\theta_m$ ($1 \leq m \leq M$) and each entry in $\mathbf{A}'$ within $[0, 1]$. 
%
where $\mu_3$ and $\mu_4$ are hyper-parameters. The second term ensures the debiased result $\mathbf{\tilde{A}}$ is close to the original structure $\mathbf{A}$. The third term enforces the debiased network structure is also sparse, which is aligned with the characteristics of real-world networks~\cite{jin2020graph}. 

\noindent \textbf{Optimization Strategy.} To optimize function $f$, parameter $\bm{\Theta}$, and $\mathbf{\tilde{A}}$, we propose a gradient-based optimization approach for alternatively training as Algorithm~\ref{algorithm} in Appendix.
First, for the optimization of $f$ w.r.t. Eq. (\ref{wd_7}), we directly utilize Stochastic Gradient Descent (SGD).
Second, for the optimization of parameter $\bm{\Theta}$ w.r.t. Eq. (\ref{wd_8}), we adopt Proximal Gradient Descent (PGD). In the projection operation in PGD, we clip the parameters in $\bm{\Theta}$ within [0, 1]. Finally, to remove the most biased attribute channels, the $z$ smallest weights in the diagonal of $\bm{\Theta}$ are masked with 0, where $z$ is a pre-assigned hyper-parameter for attribute debiasing.
Third, for the optimization of parameter $\mathbf{\tilde{A}}$ w.r.t. Eq. (\ref{wd_9}), we also adopt PGD with similar clipping strategy as the optimization of $\bm{\Theta}$. 
Finally, Algorithm~\ref{algorithm} outputs $\mathbf{\tilde{X}}$ and $\mathbf{\tilde{A}}$ after multiple epochs of optimization.

\noindent \textbf{Edge Binarization.} Here we introduce how we binarize the elements in $\mathbf{\tilde{A}}$ to indicate existence of edges. The basic intuition is to set a numerical threshold to determine the edge existence based on the entry-wise value change between $\mathbf{\tilde{A}}$ and $\mathbf{A}$. Specifically, for the "0" entries in $\mathbf{A}$, if the corresponding weight of any entry in $\mathbf{\tilde{A}}$ exceeds $r \cdot \max (\mathbf{\tilde{A}} - \mathbf{A})$, then we flip such entry from 0 to 1. Here $r$ is a pre-set threshold for binarization, and $\max (\cdot)$ outputs the largest entry of a matrix. Similarly, for the "1" entries in $\mathbf{A}$, if the corresponding weight of any entry in $\mathbf{\tilde{A}}$ is reduced by a number exceeding $r \cdot | \min (\mathbf{\tilde{A}} - \mathbf{A})|$, then such entry should be flipped as 0. Here $\min (\cdot)$ gives the smallest entry of a matrix.
To summarize, this operation aims to flip the entries with significant changes in value directly, and maintain other entries as their original values.
Finally, the binarized matrix is assigned to $\mathbf{\tilde{A}}$ as the final outcome.

\section{Experimental Evaluations}

% In this section, we perform experiments on both real-world and synthetic datasets to evaluate the effectiveness of EDITS on debiasing and our proposed two bias metrics on measuring bias in network data. In particular, our goal here is to answer the following research questions.

We conduct experiments on both real-world and synthetic datasets to evaluate the effectiveness of EDITS. In particular, we answer the following research questions.
\textbf{RQ1:} How well can EDITS mitigate the bias in attributed networks together with the outcome of different GNN variants for the downstream task?
\textbf{RQ2:} How well can EDITS balance utility maximization and bias mitigation compared with other debiasing baselines tailored for a specific GNN?
% \textbf{RQ2:} How well does the proposed two bias metrics (i.e., \textit{attribute bias} and \textit{structural bias}) consist with the bias measurements in downstream task predictions?
%  on a given GNN model

% In this section, we conduct experiments to evaluate the eectiveness and eciency of the proposed DANE framework for dynamic
% aributed network embedding. In particular, we aempt to answer
% the following two questions: (1) Eectiveness: how eective are
% the embeddings obtained by DANE on dierent learning tasks? (2)
% Eciency: how fast is the proposed framework DANE compared
% with other oine embedding methods? We rst introduce the
% datasets and experimental seings before presenting details of the
% experimental results.

% rq1 我们的指标能不能反映下游任务中的bias？
% rq2 我们的framework在有效性上如何？包括优化我们自己的指标、优化传统指标和比别人好多少
% ablation study 每个模块的有效性如何

% In this section, the adopted downstream task and adopted datasets are first presented, followed by details on experimental settings and implementation. Finally, we explore the effectiveness of EDITS based on three research questions together with the ablation study.

\subsection{Downstream Task and Datasets}
\label{data}
\textbf{Downstream Task.} We choose the widely adopted \textit{node classification} task to assess the effectiveness of our proposed framework. 

%Node classification is a common task, and it is with high practical values in various real-world scenarios~\cite{bhagat2011node,kipf2016semi,velivckovic2017graph}.

\noindent \textbf{Datasets.} We use two types of datasets in our experiments, including six real-world datasets and two synthetic datasets. Statistics of the real-world datasets can be found in Table~\ref{datasets} of Appendix. We elaborate more details as follows: (1) \textit{Real-world Datasets.} We use six real-world datasets, namely Pokec-z, Pokec-n~\cite{takac2012data,dai2020say}, UCSD34~\cite{traud2012social}, German Credit, Credit Defaulter, and Recidivism~\cite{agarwal2021towards}. 
We first introduce the three web-related networks.
\textit{Pokec-z} and \textit{Pokec-n} are collected from a popular social network in Slovakia. Here a node represents a user, and an edge denotes the friendship relation between two users~\cite{takac2012data}. We take "region" as the sensitive attribute, and the task is to predict the user working field. 
UCSD34 is a Facebook friendship network of the University of California San Diego~\cite{traud2012social}. Each node denotes a user, and edges represent the friendship relations between nodes. We take "gender" as the sensitive attribute, and the task is to predict whether a user belongs to a specific major. Users with incomplete information (e.g., missing attribute values) are filtered out from the three web networks above. Besides, we also adopt three networks beyond web-related data.
In \textit{German Credit}, nodes represent clients in a German bank, and edges are formed between clients if their credit accounts are similar. With "gender" being the sensitive attribute, the task is to classify the credit risk of the clients as high or low. 
In \textit{Recidivism}, nodes are defendants released on bail during 1990-2009. Nodes are connected based on the similarity of past criminal records and demographics. The task is to classify defendants into bail vs. no bail, with "race" being the sensitive attribute. 
In the \textit{Credit Defaulter}, nodes are credit card users, and they are connected based on the pattern similarity of their purchases and payments. Here "age" is the sensitive attribute, and the task is to predict whether a user will default on credit card payment. 
(2)  \textit{Synthetic Datasets.} For the ablation study of EDITS, we use the two datasets generated in Sec.~\ref{investigation}. One network has biased attributes and an unbiased structure, while the other network is on the opposite. We add eight extra attribute dimensions besides the two attribute dimensions for both datasets. The attribute values in the extra attribute dimensions are generated uniformly between 0 and 1. For labels, we compute the sum of the first two extra attribute dimensions. Then, we add Gaussian noise to the sum values, and rank them by the values in descending order. Labels of the top-ranked 50\% individuals are set as 1, while the labels of the other 50\% are set as 0. The task is to predict the labels.

\subsection{Experimental settings}

\noindent \textbf{GNN Models.} Here we adopt three popular GNN variants in our experiments: GCN~\cite{kipf2016semi}, GraphSAGE~\cite{hamilton2017inductive}, and GIN~\cite{xu2018powerful}.

\noindent \textbf{Baselines.} Since there is no existing work directly debiasing network data for GNNs, here we choose two state-of-the-art GNN-based debiasing approaches for comparison, namely FairGNN~\cite{dai2020say} and NIFTY~\cite{agarwal2021towards}. (1) \textit{FairGNN}. It is a debiasing method based on adversarial training. A discriminator is trained to distinguish the representations between different demographic groups. The goal of FairGNN is to train a GNN that fools the discriminator for bias mitigation. (2) \textit{NIFTY}. It is a recently proposed GNN-based debiasing framework. With counterfactual perturbation on the sensitive attribute, bias is mitigated via learning node representations that are invariant to the sensitive attribute.
It should be noted that both of them take GNNs as their backbones in the downstream task. While on the other hand, EDITS attempts at debiasing attributed networks without referring to the output of downstream GNN models (i.e., EDITS is model-agnostic). The hyper-parameters of EDITS are tuned only based on our proposed bias metrics. Obviously, the debiasing performed by EDITS is with better generalization ability but more difficult compared with the model-oriented baselines.

% \begin{itemize}
%     \item \textbf{FairGNN:} It is a debiasing method based on adversarial training. A discriminator is trained to distinguish the representations between demographic groups. The goal of FairGNN is to train a GNN that fools the discriminator for bias mitigation.
    
%     \item \textbf{NIFTY:} It is a recently proposed GNN-based debiasing framework. With counterfactual perturbation on the sensitive attribute, bias is mitigated via learning node representations that are invariant to the sensitive attribute.
% \end{itemize}

\noindent \textbf{Evaluation Metrics.} We evaluate model performance from two perspectives: model utility and bias mitigation. Good performance means low bias and high model utility. We introduce the adopted metrics for model utility and bias mitigation: (1) \textit{Model Utility Metrics}. For node classification, we use the area under the receiver operating characteristic curve (AUC) and F1 score as the indicator of model utility; (2) \textit{Bias Mitigation Metrics}. We use two widely-adopted metrics $\Delta_{SP}$ and $\Delta_{EO}$ to show to what extent the bias in the output of different GNNs are mitigated~\cite{louizos2015variational,beutel2017data,dai2020say}. For both metrics, a lower value means better bias mitigation performance.

\begin{table}[]
\vspace{-2mm}
\scriptsize
% \scriptsize
\centering
% \caption{Attribute and structural bias comparison between original networks and debiased ones from EDITS. Lower value for both metrics indicates better bias mitigation. All values are in scale of \textbf{$\times 10^{-3}$}. Best ones are marked in bold.}
\caption{Attribute and structural bias comparison between original networks and debiased ones from EDITS (in scale of \textbf{$\times 10^{-3}$}). The lower, the better. Best ones are marked in bold.}
\vspace{-3mm}
\label{bias_results}
\begin{tabular}{cccccc}
\hline
       & \multicolumn{2}{c}{\textbf{Attribute Bias}} &  & \multicolumn{2}{c}{\textbf{Structural Bias}} \\
       \cline{2-3}  \cline{5-6}
       & \textbf{Vanilla}   & \textbf{EDITS}   &  & \textbf{Vanilla}   & \textbf{EDITS}  \\
       \hline
\textbf{Pokec-z} & 0.43             &\textbf{0.33} ($-23.3\%$)             &  &0.83  & \textbf{0.75} ($-9.64\%$)           \\
\hline
\textbf{Pokec-n} & 0.54             &\textbf{0.42} ($-22.2\%$)             &  &1.03 & \textbf{0.89}  ($-13.6\%$)          \\
\hline
\textbf{UCSD34} & 0.53             &\textbf{0.48} ($-9.43\%$)             &  &0.68 & \textbf{0.63}  ($-7.35\%$)          \\
\hline
\textbf{German} & 6.33             &\textbf{2.38} ($-62.4\%$)             &  &10.4  & \textbf{3.54} ($-66.0\%$)           \\
\hline
\textbf{Credit} & 2.46             &\textbf{0.56} ($-77.2\%$)             &  &4.45 & \textbf{2.36}  ($-47.0\%$)          \\
\hline
\textbf{Recidivism}   & 0.95       &\textbf{0.39} ($-58.9\%$)             &  &1.10 & \textbf{0.52}  ($-52.7\%$)          \\
\hline
\end{tabular}
\vspace{-2.4em}
\end{table}

\begin{table*}[]
\vspace{-3mm}
\centering
\caption{Comparison on utility and bias mitigation between GNNs with original networks (denoted as Vanilla) and debiased networks (denoted as EDITS) as input. $\uparrow$ denotes the larger, the better; $\downarrow$ denotes the opposite. Best ones are in \textbf{bold}.}
\vspace{-2mm}
\label{results}
% \large
\footnotesize
% \scriptsize
\begin{tabular}{crcccccccc}
\hline
\multicolumn{2}{c}{}               & \multicolumn{2}{c}{\textbf{GCN}} &  & \multicolumn{2}{c}{\textbf{GraphSAGE}} &  & \multicolumn{2}{c}{\textbf{GIN}} \\
                                    \cline{3-4}                  \cline{6-7}                  \cline{9-10}
\multicolumn{2}{c}{}               & \textbf{Vanilla}     & \textbf{EDITS}     &  & \textbf{Vanilla}        & \textbf{EDITS}        &  & \textbf{Vanilla}     & \textbf{EDITS}     \\
\hline
\multirow{4}{*}{\textbf{Pokec-z}} & \textbf{AUC} $\uparrow$     &  \textbf{67.83 $\pm$ 0.7\%}   & 67.38 $\pm$ 0.3\%           &  & \textbf{68.00 $\pm$ 0.3\%}    & 66.37 $\pm$ 0.7\%     &  & \textbf{66.74 $\pm$ 0.8\%}            & 65.64 $\pm$ 0.5\%           \\
                        \cline{2-10}
                        & \textbf{F1} $\uparrow$                &  \textbf{61.95 $\pm$ 0.6\%}   & 61.91 $\pm$ 0.1\%           &  & \textbf{61.58 $\pm$ 1.3\%}    & 60.62 $\pm$ 0.6\%     &  & \textbf{61.55 $\pm$ 0.5\%}            & 60.65 $\pm$ 1.2\%           \\
                        \cline{2-10}
                        & $\bm{\Delta_{SP}}$ $\downarrow$       &  5.70 $\pm$ 1.2\%   & \textbf{2.74 $\pm$ 0.9\%}           &  & 7.10 $\pm$ 1.2\%                & \textbf{2.89 $\pm$ 0.4\%}     &  & 5.20 $\pm$ 1.0\%            & \textbf{1.90 $\pm$ 1.3\%}           \\
                        \cline{2-10}
                        & $\bm{\Delta_{EO}}$ $\downarrow$       &  4.88 $\pm$ 1.3\%   & \textbf{2.87 $\pm$ 1.0\%}           &  & 6.37 $\pm$ 0.8\%                & \textbf{2.54 $\pm$ 0.7\%}     &  & 4.65 $\pm$ 1.1\%             & \textbf{2.09 $\pm$ 1.1\%}           \\
                        \hline
\multirow{4}{*}{\textbf{Pokec-n}} & \textbf{AUC} $\uparrow$     &  \textbf{63.24 $\pm$ 0.5\%}   & 61.82 $\pm$ 0.9\%           &  & \textbf{64.07 $\pm$ 0.4\%}    & 62.05 $\pm$ 0.6\%     &  & \textbf{62.53 $\pm$ 1.4\%}            & 61.60 $\pm$ 1.4\%           \\
                        \cline{2-10}
                        & \textbf{F1} $\uparrow$                &  \textbf{54.32 $\pm$ 0.4\%}   & 52.84 $\pm$ 0.3\%           &  & \textbf{53.45 $\pm$ 1.2\%}     & 52.53 $\pm$ 0.1\%     &  & \textbf{52.62 $\pm$ 1.2\%}            & 52.56 $\pm$ 1.0\%           \\
                        \cline{2-10}
                        & $\bm{\Delta_{SP}}$ $\downarrow$       &  3.36 $\pm$ 0.4\%   & \textbf{0.91 $\pm$ 0.87\%}           &  & 3.85 $\pm$ 0.2\%                & \textbf{2.08 $\pm$ 1.2\%}     &  & 5.90 $\pm$ 2.5\%            & \textbf{0.96 $\pm$ 0.5\%}           \\
                        \cline{2-10}
                        & $\bm{\Delta_{EO}}$ $\downarrow$       &  3.97 $\pm$ 1.6\%   & \textbf{1.10 $\pm$ 1.0\%}           &  & 2.64 $\pm$ 0.3\%                & \textbf{1.82 $\pm$ 0.9\%}     &  & 4.47 $\pm$ 3.7\%             & \textbf{0.47 $\pm$ 0.4\%}           \\
                        \hline
\multirow{4}{*}{\textbf{UCSD34}} & \textbf{AUC} $\uparrow$     &  \textbf{63.33 $\pm$ 0.3\%}   & 62.43 $\pm$ 0.9\%           &  & 62.62 $\pm$ 1.0\%    & \textbf{62.82 $\pm$ 2.4\%}     &  & 62.57 $\pm$ 0.7\%            & \textbf{64.50 $\pm$ 0.9\%}           \\
                        \cline{2-10}
                        & \textbf{F1} $\uparrow$                &  94.16 $\pm$ 0.3\%   & \textbf{94.69 $\pm$ 0.1\%}           &  & 94.00 $\pm$ 0.2\%     & \textbf{94.55 $\pm$ 0.1\%}     &  & 92.24 $\pm$ 1.6\%            & \textbf{92.48 $\pm$ 0.5\%}           \\
                        \cline{2-10}
                        & $\bm{\Delta_{SP}}$ $\downarrow$       &  1.27 $\pm$ 0.4\%   & \textbf{0.27 $\pm$ 0.1\%}           &  & 1.27 $\pm$ 0.5\%                & \textbf{0.35 $\pm$ 0.3\%}     &  & 2.11 $\pm$ 1.3\%            & \textbf{0.36 $\pm$ 0.1\%}           \\
                        \cline{2-10}
                        & $\bm{\Delta_{EO}}$ $\downarrow$       &  1.40 $\pm$ 0.4\%   & \textbf{0.39 $\pm$ 0.1\%}           &  & 1.40 $\pm$ 0.4\%                & \textbf{0.25 $\pm$ 0.3\%}     &  & 2.32 $\pm$ 1.6\%             & \textbf{0.47 $\pm$ 0.4\%}           \\
                        \hline
\multirow{4}{*}{\textbf{German}} & \textbf{AUC} $\uparrow$      &  \textbf{74.46 $\pm$ 0.7\%}   & 71.01 $\pm$ 1.3\%           &  & \textbf{75.28 $\pm$ 2.1\%}               & 73.21 $\pm$ 0.5\%      &  & 71.35 $\pm$ 1.7\%             & \textbf{71.51 $\pm$ 0.6\%}           \\
                        \cline{2-10}
                        & \textbf{F1} $\uparrow$               &  81.54 $\pm$ 0.9\%   & \textbf{82.43 $\pm$ 0.69\%}           &  & \textbf{81.52 $\pm$ 1.0\%}               & 80.62 $\pm$ 1.5\%      &  & 83.08 $\pm$ 0.9\%             & \textbf{83.78 $\pm$ 0.4\%}           \\
                        \cline{2-10}
                        & $\bm{\Delta_{SP}}$ $\downarrow$        &  43.14 $\pm$ 2.5\%   & \textbf{2.04 $\pm$ 1.3\%}           &  & 26.83 $\pm$ 0.5\%                & \textbf{8.30 $\pm$ 3.1\%}      &  & 18.55 $\pm$ 2.0\%             & \textbf{1.26 $\pm$ 0.7\%}           \\
                        \cline{2-10}
                        & $\bm{\Delta_{EO}}$ $\downarrow$        &  33.75 $\pm$ 0.4\%   & \textbf{0.63 $\pm$ 0.39\%}           &  & 20.66 $\pm$ 3.0\%                & \textbf{3.75 $\pm$ 3.3\%}      &  & 11.27 $\pm$ 3.5\%             & \textbf{2.87 $\pm$ 1.4\%}           \\
                        % \cline{2-10}
                        % & \textbf{$b_{attr}$}        & $6.33 \times 10^{-3}$&                             &  &$6.33 \times 10^{-3}$             &                        &  & $6.33 \times 10^{-3}$         &           \\
                        % \cline{2-10}
                        % & \textbf{$b_{struc}$}       & $10.4 \times 10^{-3}$&                             &  &$10.4 \times 10^{-3}$             &                        &  & $10.4 \times 10^{-3}$         &           \\
                        \hline
\multirow{4}{*}{\textbf{Credit}} & \textbf{AUC} $\uparrow$     &  \textbf{73.62 $\pm$ 0.3\%}   & 70.16 $\pm$ 0.6\%           &  & 74.99 $\pm$ 0.2\%               & \textbf{75.28 $\pm$ 0.5\%}     &  & \textbf{73.82 $\pm$ 0.4\%}            & 72.06 $\pm$ 0.9\%           \\
                        \cline{2-10}
                        & \textbf{F1} $\uparrow$               &  \textbf{81.86 $\pm$ 0.1\%}   & 81.44 $\pm$ 0.2\%           &  & 82.31 $\pm$ 0.7\%               & \textbf{83.39 $\pm$ 0.3\%}     &  & 82.11 $\pm$ 0.1\%            & \textbf{85.10 $\pm$ 0.7\%}           \\
                        \cline{2-10}
                        & $\bm{\Delta_{SP}}$ $\downarrow$        &  12.93 $\pm$ 0.1\%   & \textbf{9.13 $\pm$ 1.2\%}           &  & 17.03 $\pm$ 3.3\%                & \textbf{12.25 $\pm$ 0.2\%}     &  & 12.18 $\pm$ 0.3\%            & \textbf{8.79 $\pm$ 5.6\%}           \\
                        \cline{2-10}
                        & $\bm{\Delta_{EO}}$ $\downarrow$        &  10.65 $\pm$ 0.0\%   & \textbf{7.88 $\pm$ 1.0\%}           &  & 15.31 $\pm$ 4.0\%                & \textbf{9.58 $\pm$ 0.1\%}     &  & 9.48 $\pm$ 0.3\%             & \textbf{7.19 $\pm$ 3.8\%}           \\
                        % \cline{2-10}
                        % & \textbf{$b_{attr}$}        & $1.78 \times 10^{-3}$&                             &  &$1.78 \times 10^{-3}$             &                        &  & $1.78 \times 10^{-3}$         &           \\
                        % \cline{2-10}
                        % & \textbf{$b_{struc}$}       & $3.34 \times 10^{-3}$&                             &  &$3.34 \times 10^{-3}$             &                        &  & $3.34 \times 10^{-3}$         &           \\
                        \hline
\multirow{4}{*}{\textbf{Recidivism}}   & \textbf{AUC} $\uparrow$ &  \textbf{86.91 $\pm$ 0.4\%}    & 85.96 $\pm$ 0.3\%           &  & 88.12 $\pm$ 1.4\%              & \textbf{88.15 $\pm$ 0.9\%}    &  & \textbf{82.40 $\pm$ 0.8\%}            & 81.55 $\pm$ 1.5\%           \\
                        \cline{2-10}
                        & \textbf{F1} $\uparrow$               &  \textbf{78.30 $\pm$ 1.0\%}    & 75.80 $\pm$ 0.5\%           &  & 76.23 $\pm$ 2.8\%              & \textbf{76.30 $\pm$ 1.4\%}    &  & 70.36 $\pm$ 1.9\%            & \textbf{71.09 $\pm$ 2.3\%}           \\
                        \cline{2-10}
                        & $\bm{\Delta_{SP}}$ $\downarrow$        &  7.89 $\pm$ 0.3\%    & \textbf{5.39 $\pm$ 0.2\%}           &  & 2.42 $\pm$ 1.2\%                 & \textbf{0.79 $\pm$ 0.5\%}    &  & 9.97 $\pm$ 0.7\%             & \textbf{4.98 $\pm$ 0.9\%}           \\
                        \cline{2-10}
                        & $\bm{\Delta_{EO}}$ $\downarrow$        &  5.58 $\pm$ 0.2\%    & \textbf{3.36 $\pm$ 0.3\%}           &  & 2.98 $\pm$ 2.2\%                 & \textbf{1.01 $\pm$ 0.5\%}    &  & 6.10 $\pm$ 1.2\%             & \textbf{5.47 $\pm$ 0.7\%}           \\
                        % \cline{2-10}
                        % & \textbf{$b_{attr}$}        & $0.95 \times 10^{-3}$&                             &  &$0.95 \times 10^{-3}$             &                        &  & $0.95 \times 10^{-3}$         &           \\
                        % \cline{2-10}
                        % & \textbf{$b_{struc}$}       & $1.10 \times 10^{-3}$&                             &  &$1.10 \times 10^{-3}$             &                        &  & $1.10 \times 10^{-3}$         &           \\
                        \hline
\end{tabular}
\vspace{-1.9em}
\end{table*}

\subsection{Debiasing Attributed Network for GNNs}
To answer \textbf{RQ1}, we first evaluate the effectiveness of EDITS in reducing the bias measured by the two proposed metrics and traditional bias metrics with different GNN backbones. The attribute and structural bias of the six real-world datasets before and after being debiased by EDITS are shown in Table~\ref{bias_results}.
The comparison on $\Delta_{SP}$ and $\Delta_{EO}$ between GNNs trained on debiased networks from EDITS and original networks is presented in Table~\ref{results}. We make the following observations: (1) From the perspective of bias mitigation in the attributed network, EDITS demonstrates significant advantages over the vanilla approach as indicated by Table~\ref{bias_results}. This verifies the effectiveness of EDITS in reducing the bias existing in the attributed network data. (2) From the perspective of bias mitigation in the downstream task, we observe from Table~\ref{results} that EDITS achieves desirable bias mitigation performance with little utility sacrifice in all cases compared with GNNs with the original network as input (i.e., the vanilla one). This verifies that attributed networks debiased by EDITS can generally mitigate the bias in the outcome of different GNNs. (3) When comparing bias mitigation performance indicated by Table~\ref{bias_results} and Table~\ref{results}, we can find that the bias in the outcome of GNNs is also mitigated after EDITS mitigates attribute bias and structural bias in the attributed networks. Such consistency verifies the validity of our proposed metrics on measuring the bias that existed in the attributed networks.
% \begin{itemize}[topsep=0pt]
%     \item From the perspective of bias mitigation in the attributed network (i.e., attribute bias and structural bias), EDITS demonstrates significant advantages over the Vanilla approach as indicated by Table~\ref{bias_results}. This verifies the effectiveness of EDITS in reducing the bias existing in the attributed network data.

%     \item From the perspective of bias mitigation in the downstream task (i.e., $\Delta_{SP}$ and $\Delta_{EO}$), we observe from Table~\ref{results} that EDITS achieves desirable bias mitigation performance with little utility sacrifice in all cases compared with GNNs with the original network as input (i.e., the vanilla one). This verifies that attributed networks debiased by EDITS can generally mitigate the bias in the outcome of different GNNs.
    
%     \item When comparing bias mitigation performance indicated by Table~\ref{bias_results} and Table~\ref{results}, we can find that the bias in the outcome of GNNs is also mitigated after EDITS mitigates attribute bias and structural bias in the attributed networks. Such consistency verifies the validity of our proposed metrics on measuring the bias that existed in the attributed networks.
% \end{itemize}

\subsection{Comparison with Other Debiasing Models}
\begin{figure}[t]
\centering
        % \vspace{-1mm}
        \begin{subfigure}[t]{0.236\textwidth}
        \setlength\figureheight{1.1in}
        \setlength\figurewidth{3.0in}
        \small
        \input{images/auc1.tikz}
        \vspace{-3mm}
            \caption[Network2]%
            {{\small AUC on web networks}}    
            \label{baseline_auc1}
            % \hspace{5pt}    % some space between two figures
        \end{subfigure}
        \centering
        \begin{subfigure}[t]{0.236\textwidth}
        \setlength\figureheight{1.1in}
        \setlength\figurewidth{3.0in}
        \small
        \input{images/auc2.tikz}
        \vspace{-3mm}
            \caption[Network2]%
            {{\small AUC on other networks}}    
            \label{baseline_auc2}
            % \hspace{5pt}    % some space between two figures
        \end{subfigure}
        \\
        % \hspace{-7pt}
        \begin{subfigure}[t]{0.23\textwidth}
        \setlength\figureheight{1.1in}
        \setlength\figurewidth{3.0in}
        \small
        \input{images/sp1.tikz} 
                \vspace{-3mm}
            \caption[Network2]%
            {{\small $\bm{\Delta_{SP}}$ on web networks}}    
            \label{baseline_sp1}
        \end{subfigure}
                \begin{subfigure}[t]{0.23\textwidth}
        \setlength\figureheight{1.1in}
        \setlength\figurewidth{3.0in}
        \small
        \input{images/sp2.tikz} 
                \vspace{-3mm}
            \caption[Network2]%
            {{\small $\bm{\Delta_{SP}}$ on other networks}}    
            \label{baseline_sp2}
        \end{subfigure}
        \vspace{-4mm}
        \caption{Performance comparison between EDITS and baselines on utility (AUC) and bias mitigation ($\bm{\Delta_{SP}}$).} 
        \vspace{-8mm}
        \label{baseline_both}
\end{figure}
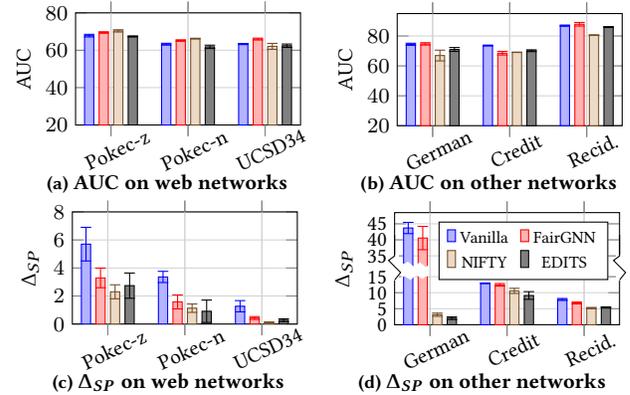

To answer \textbf{RQ2}, we then compare the balance between model utility and bias mitigation with other baselines based on a given GNN. Here we present the comparison of AUC and $\Delta_{SP}$ based on GCN in Fig.~(\ref{baseline_both}). Similar results can be obtained for other GNNs, which are omitted due to space limit. Experimental results include the performance of baselines and EDITS on the six real-world datasets. The following observations can be made: (1) From the perspective of model utility (indicated by Fig.~(\ref{baseline_auc1}) and Fig.~(\ref{baseline_auc2})), EDITS and baselines achieve comparable results with the vanilla GCN. This implies that the debiasing process of EDITS preserves as much useful information for the downstream task as the original attributed network. (2) From the perspective of bias mitigation (indicated by Fig.~(\ref{baseline_sp1}) and Fig.~(\ref{baseline_sp2})), all baselines achieve effective bias mitigation. Compared with debiasing in downstream tasks, debiasing the attributed network is more difficult due to the lack of supervision signals from GNN prediction. Observation can be drawn that the debiasing performance of EDITS is similar to or even better than that of the adopted baselines. This verifies the superior performance of EDITS on debiasing attributed networks for more fair GNNs. (3) From the perspective of balancing the model utility and bias mitigation, EDITS achieves comparable model utility with alternatives but exhibits better bias mitigation performance. Consequently, we argue that EDITS achieves superior performance on balancing the model utility and bias mitigation over other baselines.

% \begin{figure}[!htbp]
%         % \centering
%                 % \vspace{-2mm}
%         \begin{subfigure}[t]{0.43\textwidth}
%         \setlength\figureheight{1.45in}
%         \setlength\figurewidth{3.0in}
%         % \centering  \scriptsize
%         \small
%         \input{images/auc_baseline.tikz}
%             \caption[Network2]%
%             {{\small Comparison on AUC}}    
%             \label{baseline_auc}
%             \hspace{5pt}    % some space between two figures
%         \end{subfigure}
%         \hfill
%         \begin{subfigure}[t]{0.456\textwidth}
%         % \setlength\figureheight{1.65in}
%         % \setlength\figurewidth{3.2in}
%         \setlength\figureheight{1.45in}
%         \setlength\figurewidth{3.0in}
%         % \centering  \scriptsize
%         \small
%         \input{images/example_broken.tikz} 
%             \caption[Network2]%
%             {{\small Comparison on $\bm{\Delta_{SP}}$}}    
%             \label{baseline_sp}
%         \end{subfigure}
%         \vspace{-2mm}
%         \caption{Performance comparison between EDITS and baselines on utility (AUC) and bias mitigation ($\bm{\Delta_{SP}}$) on GCN.} 
%         \vspace{-2mm}
%         \label{baseline_both}
% \end{figure}

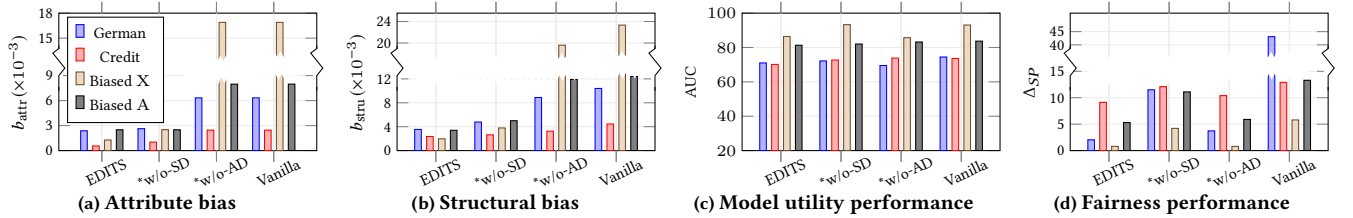
\begin{figure*}[!htbp]
\vspace{-2.5mm}
        \centering
        \begin{subfigure}[t]{0.23\textwidth}  % 0.23
        \setlength\figureheight{0.9in}
        \setlength\figurewidth{1.35in}
        \centering  \scriptsize
        \input{images/to_bar_attr.tikz}
                    \vspace{-5mm}
            \caption[Network2]%
            {{\small Attribute bias}}    
            \label{ablation1}
        \end{subfigure}
        \hspace{7pt}
        \begin{subfigure}[t]{0.23\textwidth}
        \setlength\figureheight{0.9in}
        \setlength\figurewidth{1.35in}
        \centering  \scriptsize
        \input{images/to_bar_stru.tikz}
                            \vspace{-5mm}
            \caption[Network2]%
            {{\small Structural bias}}    
            \label{ablation2}
        \end{subfigure}
        \hspace{7pt}
        \centering
        \begin{subfigure}[t]{0.23\textwidth}
        \setlength\figureheight{0.9in}
        \setlength\figurewidth{1.35in}
        \centering  \scriptsize
        \input{images/to_bar_auc.tikz} 
                                    \vspace{-5mm}
            \caption[Network2]%
            {{\small Model utility performance}}    
            \label{ablation3}
        \end{subfigure}
        \hspace{7pt}
        \begin{subfigure}[t]{0.23\textwidth}
        \setlength\figureheight{0.9in}
        \setlength\figurewidth{1.35in}
        \centering  \scriptsize
        \input{images/to_bar_sp.tikz} 
                                    \vspace{-5mm}
            \caption[Network2]%
            {{\small Fairness performance}}    
            \label{ablation4}
        \end{subfigure}
        % There must be a [] after the \caption command.
        \vspace{-4mm}
        \caption[]{Performance EDITS and its variants on two real-world datasets and two synthetic datasets. EDITS denotes that both debiasing modules are included; *w/o-SD means EDITS without structural debiasing module; *w/o-AD means EDITS is without attribute debiasing module; Vanilla means applying GNN with the original attributed network as input.}  % \examplelegend:  
        \vspace{-4.9mm}
        \label{ablation}
\end{figure*}

\subsection{Ablation Study}
To evaluate the effectiveness of the two debiasing modules (i.e., attribute debiasing module and structural debiasing module) in EDITS, here we investigate how each of them individually contributes to bias mitigation under our proposed bias metrics and the traditional bias metrics in the downstream task. We choose GCN as the GNN model in our downstream task. For better visualization purposes, the two datasets showing large attribute bias and structural bias (i.e., \textit{German} and \textit{Credit}) are selected for experiments. Besides, to better demonstrate the functionality of the two debiasing modules, we also adopt the two synthetic datasets we mentioned in Sec.~\ref{investigation} (i.e., the network with only biased attributes and the network with only biased structure), which are further modified according to Sec.~\ref{data}. Based on the four selected datasets, four different variants of EDITS are tested, namely EDITS with both debiasing modules, EDITS without the structural debiasing module (i.e., *w/o-SD), EDITS without the attribute debiasing module (i.e., *w/o-AD), vanilla GCN model without debiased input (i.e., Vanilla). We present their performance of attribute bias, structural bias, AUC, and $\Delta_{SP}$ on the four datasets in Fig.~(\ref{ablation}).
We make the following observations: (1) The value of $\textit{attribute bias}$ can be reduced with the attribute debiasing module of EDITS, which maintains the model utility (i.e., AUC) but reduces $\Delta_{SP}$ in the downstream task. (2) The value of $\textit{structural bias}$ can be reduced with both attribute debiasing and structural debiasing modules. With only structural debiasing, EDITS still maintains comparable model utility but reduces $\Delta_{SP}$ in the downstream task. (3) Although both attribute debiasing and structural debiasing module help mitigate $\textit{structural bias}$, only debiasing the network structure achieves better bias mitigation performance on all four datasets compared with only debiasing the attributes as implied by Fig.~(\ref{ablation4}). This demonstrates the indispensability of the structural debiasing module in EDITS.
% \begin{itemize}[topsep=0pt]
%     \item The value of $\textit{attribute bias}$ can be reduced with the attribute debiasing module of EDITS, which maintains the model utility (i.e., AUC) but reduces $\Delta_{SP}$ in the downstream task.
%     \item The value of $\textit{structural bias}$ can be reduced with both attribute debiasing and structural debiasing modules. With only structural debiasing, EDITS still maintains comparable model utility but reduces $\Delta_{SP}$ in the downstream task.
%     \item Although both attribute debiasing and structural debiasing module help mitigate $\textit{structural bias}$, only debiasing the network structure achieves better bias mitigation performance on all four datasets compared with only debiasing the attributes as implied by Fig.~(\ref{ablation4}). This demonstrates the indispensability of the structural debiasing module in EDITS.
% \end{itemize}

\section{\vspace{-0.2em}Related Work}

% ~\cite{du2019fairness, DBLP:conf/nips/HardtPNS16, barocas2017fairness, wu2019counterfactual,mehrabi2019survey}

% ~\cite{mehrabi2019survey,du2019fairness}

\noindent \textbf{Mitigating Bias in Machine Learning.}
Bias can be defined from a variety of perspectives in machine learning algorithms~\cite{du2019fairness, DBLP:conf/nips/HardtPNS16, barocas2017fairness, wu2019counterfactual,mehrabi2019survey,li2021dyadic}. Commonly used algorithmic bias notions can be broadly categorized into \textit{group fairness} and \textit{individual fairness}~\cite{DBLP:conf/innovations/DworkHPRZ12}. 
Group fairness emphasizes that algorithms should not yield discriminatory outcomes for any specific demographic groups~\cite{DBLP:conf/innovations/DworkHPRZ12}. Such groups are usually determined by sensitive attributes, e.g., gender or race~\cite{kamiran2012data}. Existing debiasing approaches work in one of the three data flow stages, i.e., pre-processing, processing and post-processing stage. In pre-processing stage, a common method is to re-weight training samples from different groups to mitigate bias before model training~\cite{kamiran2012data}. Perturbing data distributions between groups is another popular approach to debias the data in the pre-processing stage~\cite{wang2019repairing}. In processing stage, a popular method is to add regularization terms to disentangle the outcome from sensitive attribute~\cite{ross2017right,liu2019incorporating} or minimize the outcome difference between groups~\cite{agarwal2018reductions}. Besides, utilizing adversarial learning to remove sensitive information from representations is also widely adopted~\cite{elazar2018adversarial}. In post-processing stage, bias in outcomes is usually mitigated by constraining the outcome to follow a less biased distribution~\cite{zhao2017men,DBLP:conf/nips/HardtPNS16,pleiss2017fairness,laclau2021all,krasanakis2020applying}. Usually, all above-mentioned approaches are evaluated via measuring how much certain fairness notion is violated. \textit{Statistical Parity}~\cite{DBLP:conf/innovations/DworkHPRZ12}, \textit{Equality of Opportunity}, \textit{Equality of Odds}~\cite{DBLP:conf/nips/HardtPNS16} and \textit{Counterfactual Fairness}~\cite{kusner2017counterfactual} are commonly studied fairness notions.
Different from group fairness, individual fairness focuses on treating similar individuals similarly~\cite{DBLP:conf/innovations/DworkHPRZ12,zemel2013learning}. The similarity can be given by oracle similarity scores from domain experts~\cite{lahoti2019operationalizing}. Most existing debiasing methods based on individual fairness work in the processing stage. For example, constraints can enforce similar predictions between similar instances~\cite{lahoti2019operationalizing,jung2019eliciting}. \textit{Consistency} is a popular metric for individual fairness evaluation~\cite{lahoti2019operationalizing,lahoti2019ifair}.

\noindent \textbf{Mitigating Bias in Graph Mining.}
Efforts have been made to mitigate bias in graph mining algorithms, where these works can be broadly categorized into either focusing on \textit{group fairness} or \textit{individual fairness}. For group fairness, adversarial learning can be adopted to learn less biased node representations that fool the discriminator~\cite{DBLP:conf/icml/BoseH19,dai2020say}. Rebalancing between groups is also a popular approach to mitigate bias~\cite{burke2017balanced,DBLP:journals/corr/abs-1809-09030,DBLP:conf/ijcai/RahmanS0019,tang2020investigating,li2021dyadic}. For example, Rahman et al. mitigate bias via rebalancing the appearance rate of minority groups in random walks~\cite{rahman2019fairwalk}.
Projecting the embeddings onto a hyperplane orthogonal to the hyperplane of sensitive attributes is another approach for bias mitigation~\cite{DBLP:journals/corr/abs-1909-11793}. 
Compared with the vast amount of works on group fairness, only few works promote individual fairness in graphs. To the best of our knowledge, Kang et al.~\cite{kang2020inform} first propose to systematically debias multiple graph mining algorithms based on individual fairness. 
Dong et al.~\cite{dong2021individual} argue that for each individual, if the similarity ranking of others in the GNN outcome follows the same order of an oracle ranking given by domain experts, then people can get a stronger sense of fairness.
Different from the above approaches, this paper proposes to directly debias the attributed networks in a model-agnostic way.

\section{\vspace{-0.2em}Conclusion}
GNNs are increasingly critical in various applications. Nevertheless, there is an increasing societal concern that GNNs could yield discriminatory decisions towards certain demographic groups. Existing debiasing approaches are mainly tailored for a specific GNN. Adapting these methods to different GNNs can be costly, as they need to be fine-tuned. Different from them, in this paper, we propose to debias the attributed network for GNNs. With analysis of the source of bias existing in different data modalities, we define two kinds of bias with corresponding metrics, and formulate a novel problem of debiasing attributed networks for GNNs. To tackle this problem, we then propose a principled framework EDITS for model-agnostic debiasing. Experiments demonstrate the effectiveness of EDITS in mitigating the bias and maintaining the model utility.

%
% The consistency between debiasing the attributed network and the bias mitigation in the outcome of GNNs also verifies the validity of our bias metrics.

\section{\vspace{-0.2em}ACKNOWLEDGMENTS}
This material is supported by the National Science Foundation (NSF) under grants $\#$2006844 and the Cisco Faculty Research Award.

\clearpage
\bibliographystyle{ACM-Reference-Format}
\bibliography{ref}

\appendix

\section{Appendix}

\subsection{Datasets Statistics}
\label{statistics}

The detailed statistics of six real-world datasets (i.e., German Credit, Recidivism and Credit Defaulter) can be found in Table~\ref{datasets}.

\begin{table*}[]
\vspace{-3mm}
\caption{The statistics and basic information about the six real-world datasets adopted for experimental evaluation. Sens. represents the semantic meaning of sensitive attribute.}
\vspace{-3mm}
\label{datasets}
\footnotesize
\centering
\begin{tabular}{lcccccc}
\hline
\textbf{Dataset}          & \textbf{Pokec-z}   & \textbf{Pokec-n}  & \textbf{UCSD34}    & \textbf{German Credit}        & \textbf{Recidivism}   & \textbf{Credit Defaulter}       \\
\hline
\textbf{\# Nodes}           & 7,659             & 6,185    & 4,132     & 1,000                      & 18,876             & 30,000                    \\
\textbf{\# Edges}              & 29,476             & 21,844    & 108,383    & 22,242                     & 321,308            & 1,436,858               \\
\textbf{\# Attributes}     & 59             & 59 & 7   & 27                         & 18                 & 13                         \\
\textbf{Avg. degree}         & 7.70             & 7.06 & 52.5  & 44.5                      & 34.0              & 95.8                      \\
\textbf{Sens.}            & Region             & Region & Gender  & Gender       & Race  & Age     \\
\textbf{Label}         & Working field             & Working field   & Student major  & Credit status & Bail decision   & Future default   \\
\hline
\end{tabular}
\vspace{-1.0em}
\end{table*}

\subsection{Algorithm}
\label{algorithm}

We present the optimization algorithm for EDITS in Algorithm~\ref{algorithm}.

\vspace{-4mm}
\begin{algorithm}[H]
\scriptsize
\caption{The Optimization Algorithm for EDITS}
\label{algorithm}
\begin{algorithmic}[1]
\REQUIRE ~~\\ %算法的输入参数：Input
$\mathbf{A}$: Adjacency matrix; $\mathbf{X}$: Attribute matrix; 
$\alpha$, $\mu_1$ to $\mu_4$: Hyper-parameters in objectives; 
$c$: Threshold enforcing Lipschitz; 
$z$: Threshold for attribute masking; 
$r$: Threshold factor for adjacency matrix binarization; 
% \textit{epoch\_max}: Maximum number of training epochs. 
\ENSURE ~~\\ %算法的输入参数：Input
Debiased adjacency matrix $\mathbf{\tilde{A}}$ and attribute matrix $\mathbf{\tilde{X}}$; \\
\STATE $\mathbf{\tilde{A}} \gets \mathbf{A}$; $\bm{\Theta} \gets \mathbf{I}$; 
% \STATE \textit{epoch} $\gets$ 0; 
\WHILE{\textit{epoch} $\leq$ \textit{epoch\_max}}
\STATE Compute $\mathscr{L}_1$ following Eq. (\ref{wd_6});
\STATE Update the weights of $f$ by SGD following Eq. (\ref{wd_7});
\STATE Clip the weights of $f$ within [-$c$, $c$];
\STATE Update $\bm{\Theta}$ by PGD following Eq. (\ref{wd_8}), $\mathbf{\tilde{X}} \gets \mathbf{X} \bm{\Theta}$;
% \STATE $\mathbf{\tilde{X}} \gets \mathbf{X} \bm{\Theta}$;
\STATE Update $\mathbf{\tilde{A}}$ by PGD following Eq. (\ref{wd_9}), $\mathbf{\tilde{A}} \gets \frac{1}{2} (\mathbf{\tilde{A}} + \mathbf{\tilde{A}}^{\top})$;
% \STATE $\mathbf{\tilde{A}} \gets \frac{1}{2} (\mathbf{\tilde{A}} + \mathbf{\tilde{A}}^{\top})$;
% \STATE \textit{epoch} $\gets$ \textit{epoch} + 1;
\ENDWHILE
% \IF{$z > \text{the number of 0's in }diag(\bm{\Theta})$}
\STATE Mask the $z$ smallest entries with 0 in $diag(\bm{\Theta})$, $\mathbf{\tilde{X}} \gets \mathbf{X} \bm{\Theta}$;
% \STATE $\mathbf{\tilde{X}} \gets \mathbf{X} \bm{\Theta}$;
\STATE Binarize $\mathbf{\tilde{A}}$ w.r.t. the threshold $r$;
% \ENDIF
\RETURN $\mathbf{\tilde{A}}$ and $\mathbf{\tilde{X}}$;
\end{algorithmic}
\end{algorithm}
% \vspace{-1mm}

\subsection{Theoretical Analysis}
\label{theo}

Here we present theoretical analysis for the two proposed metrics to gain a deeper understanding of debiasing attributed networks for GNNs.
% \subsection{Metric Analysis}
% equivalent to 衡量不同group在不同频率占据主要成分时，attribute distribution的差异
% \red{For \textit{attribute bias}, it is straightforward that if the Wasserstein distance of the attribute value distribution between the two groups is zero for every dimension, then there would be no clue to distinguish between the two groups. Consequently, we mainly focus on the theoretical analysis of the \textit{structural bias} metric.}
For attribute bias, it is straightforward that if the Wasserstein distance of the attribute value distribution between the two groups is zero for every dimension, then there would be no clue to distinguish between the two groups. Consequently, here we mainly focus on the theoretical analysis of the structural bias metric. Specifically, we perform theoretical analysis from the perspective of Spectral Graph Theory~\cite{chung1997spectral}. 
Usually, an undirected attributed network is regarded as a signal composed of different frequency components in Graph Signal Processing (GSP). %~\cite{rabiner2016theory}. 
If an operation preserves lower frequency components more than higher frequency components of a graph signal, this operation can be regarded as low-pass filtering the input graph signal. %~\cite{rabiner2016theory,nt2019revisiting}.

\textbf{Theorem 1.} \textit{Let $\lambda_{\text{max}}$ be the largest eigenvalue of $\mathbf{L}_{\text{norm}}$. Multiplying $\mathbf{X}$ by the propagation matrix $\mathbf{M}_{H}$ can be regarded as low-pass filtering $\mathbf{X}$ when $\alpha = \frac{1}{\lambda_{max}}$ and $\beta_i > 0$ ($1 \leq i \leq H$).}
%The resultant reachability matrix $\mathbf{R}$ is $\mathbf{X}_{\text{norm}}$ after being processed by the low-pass filter $\mathbf{M}_{H}$.}

\begin{proof}
We present the proof based on Laplacian graph spectrum. By replacing $\alpha$ with $\frac{1}{\lambda_{\text{max}}}$, we have
% \begin{align}
% \mathbf{P}_{\text{norm}} &= \frac{1}{\lambda_{\text{max}}} \mathbf{A}_{\text{norm}} + (1 - \frac{1}{\lambda_{\text{max}}}) \mathbf{I}   \notag \\
% &= \mathbf{D}^{- \frac{1}{2}} (\frac{1}{\lambda_{\text{max}}} \mathbf{A} + (1 - \frac{1}{\lambda_{\text{max}}}) \mathbf{D}) \mathbf{D}^{- \frac{1}{2}}   \notag \\
% &= \mathbf{I} - \frac{\mathbf{L}_{\text{norm}}}{\lambda_{\text{max}}}
% \end{align}
\begin{align}
\label{new_p}
\mathbf{P}_{\text{norm}} = \frac{1}{\lambda_{\text{max}}} \mathbf{A}_{\text{norm}} + (1 - \frac{1}{\lambda_{\text{max}}}) \mathbf{I} = \mathbf{I} - \frac{\mathbf{L}_{\text{norm}}}{\lambda_{\text{max}}} .
\end{align}
Then, by combining Eq. (\ref{eq:Mh}) and Eq. (\ref{new_p}), we get
\begin{small}
\begin{align}
\mathbf{M}_{H} = \beta_1 (\mathbf{I} - \frac{\mathbf{L}_{\text{norm}}}{\lambda_{\text{max}}}) + \beta_2 (\mathbf{I} - \frac{\mathbf{L}_{\text{norm}}}{\lambda_{\text{max}}})^2 + ... + \beta_H (\mathbf{I} - \frac{\mathbf{L}_{\text{norm}}}{\lambda_{\text{max}}})^H .
\label{l_in_proof}
\end{align}
\end{small}Considering that $\mathbf{L}_{\text{norm}}$ is a symmetric real matrix, it can be decomposed as $\mathbf{L}_{\text{norm}} = \mathbf{U} \mathbf{\Lambda} \mathbf{U}^{\top}$, then Eq. (\ref{l_in_proof}) can be rewritten as 
\begin{small}
\begin{align}
\label{frequency}
\mathbf{M}_{H} &= \mathbf{U}  \big(  \beta_1 (\mathbf{I} - \frac{\mathbf{\Lambda}}{\lambda_{\text{max}}}) + \beta_2 (\mathbf{I} - \frac{\mathbf{\Lambda}}{\lambda_{\text{max}}})^2 + ... + \beta_H (\mathbf{I} - \frac{\mathbf{\Lambda}}{\lambda_{\text{max}}})^H \big)  \mathbf{U}^{\top}.
\end{align}
\end{small}Here $\mathbf{\Lambda}$ is the diagonal eigenvalue matrix of $\mathbf{L}_{\text{norm}}$, and the $h$-th term ($1 \leq h \leq H$) in Eq. (\ref{frequency}) indicates a frequency response function of $(1 - \frac{\mathbf{\lambda_{i}}}{\lambda_{\text{max}}})^h$.
For any $\lambda_{i}$ ($1 \leq i \leq N$), $ \frac{\lambda_{i}}{\lambda_{\text{max}}} \leq 1$ holds. 
Consequently, the frequency response of each term in Eq. (\ref{frequency}) monotonically decreases w.r.t. $\lambda_i$. This indicates that, for each term, when it is multiplied by a graph signal, the higher frequency components of the graph signal are more weakened compared with the lower frequency components.
Therefore, according to Eq. (\ref{frequency}), $\mathbf{M}_{H}$ can be regarded as a graph filter whose frequency response is composed of $H$ low-pass filters.
In conclusion, multiplying the propagation matrix $\mathbf{M}_{H}$ with any graph signal equals to the operation of low-pass filtering when $\alpha = \frac{1}{\lambda_{\text{max}}}$ and all $\beta_i > 0$. The graph signal is the attribute matrix $\mathbf{X}$ in the proposed structural bias metric.
%The node attributes $\mathbf{X}_{\text{norm}}$ can be filtered by multiplying $\mathbf{M}_{H}$ as information propagation.
\end{proof}
% \in [0, \lambda_{\text{max}}]$

Based on Theorem 1, we propose the corollary below to build connections between attribute bias and structural bias.

\textbf{Corollary 1.} \textit{The attribute bias contained in the low frequency components of an attributed network is equivalent to structural bias.}

% Structural bias is equivalent to the attribute bias contained in the low frequency components of an attributed network.}

From the proof of Theorem 1, we can observe that $\mathbf{M}_{H} \mathbf{X}_{\text{norm}}$ is equivalent to low-pass filtering the attribute matrix $\mathbf{X}_{\text{norm}}$. Then Corollary 1 is self-evident based on Definition~\ref{defn:structural_bias}.
At the same time, considering that the frequencies and the corresponding basis of a network data changes when $\mathbf{A}$ is optimized to be $\mathbf{\tilde{A}}$. %~\cite{chung1997spectral}, 
% 这里不好说是全频段的debias+低频段的debias？？
The basic goal of EDITS can also be interpreted as: \textit{debiasing the full spectrum of a graph signal, and learning better frequencies together with the corresponding basis to further mitigate the bias existed in the lower frequency components of the graph signal}.

\subsection{Implementation Details}

EDITS is implemented using Pytorch~\cite{paszke2017automatic} and optimized via RMSprop optimizer~\cite{hinton2012neural}. 
In the training of EDITS, we set the training epochs as 100 for Recidivism and 500 for other datasets. The learning rate is set as $3 \times 10^{-3}$ for epochs under 400 and $1 \times 10^{-3}$ for those above. $\alpha$ is set as 0.5 considering that $\lambda_{\text{max}}=2$~\cite{chung1997spectral}.
%
% For a more stable optimization, the three modules in EDITS are updated alternatively.
%
% The attributed networks output by EDITS are used for GNN training to evaluate the performance of EDITS, while the performance of the Vanilla method is obtained by feeding GNNs with the original networks.
%
To train GNNs, we fix the training epochs to be 1,000 based on Adam optimizer~\cite{kingma2014adam}, with the learning rate of $1 \times 10^{-3}$. Dropout rate and hidden channel number is set as 0.05 and 16, respectively.
%
% All GNNs and baseline methods are optimized with Adam optimizer~\cite{kingma2014adam}.
%
% For baseline methods, the hyper-parameter setting follows the guidelines provided by the original papers.  
%
% More details can be found in our released implementations. 

\subsection{Extension to Non-Binary Sensitive Attributes}

Here, we show how our proposed framework EDITS can be generalized to handle non-binary sensitive attributes. More specifically, we use a synthetic dataset to showcase the extension.

\noindent \textbf{Synthetic Dataset Generation.}
Our goal here is to generate a synthetic attributed network with both biased node attributes and network structure, where nodes should come from at least three different groups based on the sensitive attribute.
We elaborate more details from three perspectives: biased network structure generation, biased node attribute generation, and node label generation. (1) \textit{Biased Network Structure Generation.} We adopt a similar approach as presented in Fig.~(\ref{bias_incorporating}) to generate three communities with dense intra-community links but sparse inter-community links. 
(2) \textit{Biased Node Attributes Generation.} We generate a ten-dimensional attribute vector for each node. The values at the first two dimensions are generated independently with Gaussian distribution $\mathcal{N}$(-1, $1^2$), $\mathcal{N}$(0, $1^2$), and $\mathcal{N}$(1, $1^2$) for the nodes in the three communities, respectively.
The attribute values for all other dimensions are generated with independent Gaussian Distribution $\mathcal{N}$(0, $1^2$). Besides, We generate a ternary variable $s \in \{0,1,2\}$ based on the node community membership for all nodes as an extra attribute dimension. Here the community membership is regarded as the sensitive attribute of nodes in this network. 
(3) \textit{Node Label Generation.} We sum up the values at the first two unbiased attribute dimensions for all nodes, and then add Gaussian noise to the summation values. The summation values with noise are ranked in descending order. Labels of the top-ranked 50\% nodes are set as 1, while the labels of the other 50\% nodes are set as 0. The task is to predict the generated labels.
%
% We present an intuitive visualization of the generated synthetic dataset in Fig.~\ref{ternary_visualization}.

% \begin{figure}[]
%     \centering
%     \subfloat[Biased attributes]{
%         \includegraphics[width=0.24\textwidth]{images/dis_2_1.pdf}
%         \label{ternary_attributes}
%     }
%     \subfloat[Biased structure]{
%         \includegraphics[width=0.24\textwidth]{images/st_1.pdf}
%         \label{ternary_structure}
%     }
%     \vspace{-3mm}
%     \caption{The visualization of the synthetic dataset for ternary sensitive attributes.}
%     \vspace{-5mm}
%     \label{ternary_visualization}
% \end{figure} 

\noindent \textbf{Framework Extension.}
To extend the proposed framework EDITS to handle non-binary sensitive attributes,
the basic rationale is to encourage the function $f_m$ introduced in Section~\ref{optimization} to help approximate the squared Wasserstein distance sum between all group pairs based on ternary sensitive attribute. Therefore, we modify the $\mathscr{L}_1$ in Eq.~(\ref{wd_6}) as
\begin{align}
\label{non-binary-l1}
\mathscr{\tilde{L}}_1 = \sum_{i,j}  \sum_{m} \{ \mathbb{E}_{\mathbf{x}_{i,m}}&[f_{m}(\mathbf{x}_{i,m})] - \mathbb{E}_{\mathbf{x}_{j,m}}[f_{m}(\mathbf{x}_{j,m})] \}^2.
\end{align}
Here $1 \leq m \leq M$, and $ i,j \in \{0,1,2\}$ ($i < j$). $\mathbf{x}_{i,m}$ and $\mathbf{x}_{j,m}$ follows $P_{i,m}^{Joint}$ and $P_{j,m}^{Joint}$, respectively.
The $\mathscr{L}_1$ in Eq.~(\ref{wd_7}),~(\ref{wd_8}), and~(\ref{wd_9}) are repalced with $\mathscr{\tilde{L}}_1$. This enables EDITS to approximate and minimize the squared Wasserstein distance sum between all group pairs.

% Research Questions \& 
\noindent \textbf{Research Questions.}
Here we aim to answer two research questions. \textbf{RQ1:} Can EDITS mitigate the bias in the network dataset with ternary sensitive attributes? \textbf{RQ2:} Can EDITS achieve a good balance between mitigating bias and maintaining utility for GNN predictions with ternary sensitive attributes?

\noindent \textbf{Evaluation Metrics.}
We introduce the metrics following the two research questions above. (1) For RQ1, to measure the bias in the network dataset, we adopt the $b_{\text{attr}}$ and $b_{\text{stru}}$ introduced in Sec.~\ref{metrics}. (2) For RQ2, to measure the bias exhibited in GNN predictions, we adopt two traditional fairness metrics: $\Delta_{SP}$ and $\Delta_{EO}$. Considering that these two metrics are designed only for binary sensitive attributes, $\Delta_{SP}$ and $\Delta_{EO}$ for each pair of groups are utilized to evaluate the fairness level of GNN predictions. Besides, AUC and F1 are adopted to evaluate the utility of GNN predictions.
%
% The fairness level is compared before and after the dataset is debiased with EDITS for both the synthetic dataset and GNN prediction. To reflect the fairness level of the synthetic dataset, we present the Attribute Bias and Structural Bias for every pair of groups based on the sensitive attribute.
% %
% We also present the $\Delta_{SP}$ and $\Delta_{EO}$ for every pair of groups as the evaluation of fairness for GNN output, which is consistent with traditional fairness metrics.
% %
% Results based on GCN are presented here, and similar observations can also be found on other GNN backbones.

\begin{figure}[t]
        \begin{subfigure}[t]{0.24\textwidth}
        \setlength\figureheight{1.1in}
        \setlength\figurewidth{3.0in}
        \small
        \input{images/non_binary_sp.tikz}
                \vspace{-2mm}
            \caption[Network2]%
            {{\small Comparison of $\bm{\Delta_{SP}}$}}    
            \label{sp_ternary}
        \end{subfigure}
                \hspace{-7pt}
        \begin{subfigure}[t]{0.24\textwidth}
        \setlength\figureheight{1.1in}
        \setlength\figurewidth{3.0in}
        \small
        \input{images/non_binary_eo.tikz}
                \vspace{-2mm}
            \caption[Network2]%
            {{\small Comparison of $\bm{\Delta_{EO}}$}}    
            \label{eo_ternary}
        \end{subfigure}
        \vspace{-3mm}
        \caption{Comparison of $\bm{\Delta_{SP}}$ and $\bm{\Delta_{EO}}$ between vanilla and EDITS based on GCN for ternary sensitive attributes.} 
        \vspace{-1mm}
        \label{sp_eo_ternary}
\end{figure}
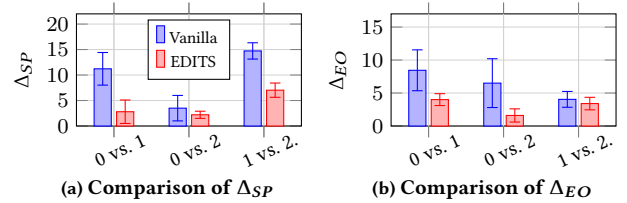

\noindent \textbf{Results Analysis.}
Results based on GCN are presented in Fig.~(\ref{sp_eo_ternary}) and Table~\ref{b_in_non_binary}, and similar observations can also be found on other GNN backbones.
We evaluate the performance of EDITS from two perspectives.
(1) \textit{RQ1: the fairness level of the network dataset}. As presented in Table~\ref{b_in_non_binary}, $b_{\text{attr}}$ and $b_{\text{stru}}$ of the dataset are clearly reduced with EDITS. This verifies the effectiveness of EDITS on debiasing the attributed network data.
(2) \textit{RQ2: the balance between fairness and utility for GNN predictions}. As presented in Fig.~(\ref{sp_eo_ternary}), $\Delta_{SP}$ and $\Delta_{EO}$ for every group pair are reduced. This corroborates the effectiveness of EDITS on achieving more fair GNN predictions. At the same time, Table~\ref{b_in_non_binary} indicates that the GNN with debiased input data still maintains similar utility performance compared with the GNN with vanilla input. This indicates that EDITS achieves a good balance between fairness and utility for GNN predictions.

\begin{table}[]
\centering
\footnotesize
\vspace{-1mm}
\caption{Parameter study for $\mu_1$ and $\mu_3$. The values of $b_{\text{attr}}$ and $b_{\text{stru}}$ are in scale of \textbf{$\times 10^{-3}$}.}
\vspace{-3mm}
\label{param_study}
\begin{tabular}{cccc||cccc}
\hline
$\mu_1$ & $b_{attr}$ & F1(\%) & $\Delta_{SP}$(\%) & $\mu_3$ & $b_{stru}$ & F1(\%) & $\Delta_{SP}$(\%) \\
\hline
 \textbf{1e2}        & 6.33           & 81.69   & 35.3              & \textbf{1e2}         & 10.2           & 82.26   & 34.2              \\
  \textbf{1e1}        & 5.02           & 80.69   & 19.9              & \textbf{1e1}         & 9.97           & 80.89   & 25.0              \\
  \textbf{1e0}        & 3.74           & 80.28   & 7.76              & \textbf{1e0}         & 9.81           & 79.77   & 14.1              \\
  \textbf{1e-1}       & 2.38           & 80.00   & 4.58              & \textbf{1e-1}        & 4.89           & 79.46   & 3.96              \\
  \textbf{1e-2}       & 2.34           & 79.95   & 4.08              & \textbf{1e-2}        & 3.53           & 78.93   & 3.26              \\
  \textbf{1e-3}       & 2.35           & 79.46   & 3.96              & \textbf{1e-3}        & 3.34           & 78.89   & 2.76              \\
  \textbf{1e-4}       & 2.34           & 79.03   & 3.29              & \textbf{1e-4}        & 3.29           & 78.37   & 2.06              \\
  \textbf{1e-5}       & 2.34           & 76.22   & 2.86              & \textbf{1e-5}        & 3.22           & 78.06   & 2.00             \\
\hline
\end{tabular}
\end{table}

\subsection{Parameter Study}

Here we aim to study the sensitivity of EDITS w.r.t. hyper-parameters. Specifically, we show the parameter study of $\mu_1$ and $\mu_3$ on German dataset, but similar observations can also be found on other datasets. Here $\mu_1$ and $\mu_3$ control how much original information should be preserved from the original attributes and graph structure, respectively. 
We first vary $\mu_1$ in the range of \{1e2, 1e1, 1e0, 1e-1, 1e-2, 1e-3, 1e-4, 1e-5\} while fix other parameters as $\mu_2$=1e-4, $\mu_3$=1e-1, $\mu_4$=1e-4; then we vary $\mu_1$ in the same range with $\mu_1$=1e-3, $\mu_2$=1e-4, $\mu_4$=1e-4.
%
% We present a parameter study based on the German dataset with GCN in Table~\ref{param_study}. Here $\mu_1$ and $\mu_3$ control how much original information should be preserved from the original attributes and graph structure, respectively. 
% %
% The study for $\mu_1$ and $\mu_3$ is carried out based on ($\mu_2$=1e-4, $\mu_3$=1e-1, $\mu_4$=1e-4) and ($\mu_1$=1e-3, $\mu_2$=1e-4, $\mu_4$=1e-4), respectively.
%
The results in Table~\ref{param_study} indicate that the trade-off between debiasing and utility performance is stable when $\mu_1$ and $\mu_3$ are in a wide range between 1e-3 and 1e-1. Therefore, it is safe to say that we can tune these parameters in a wide range without greatly affecting the fairness and model utility.

\begin{table}[]
\centering
\footnotesize
\vspace{-1mm}
\caption{Comparison of fairness level and utility between the original synthetic network and the debiased one based on the ternary sensitive attributes. The values of $b_{\text{attr}}$ and $b_{\text{stru}}$ are in scale of \textbf{$\times 10^{-3}$}. Best ones are marked in bold.}
\vspace{-3mm}
\label{b_in_non_binary}
\begin{tabular}{ccccccccc}
\hline
\hline
\multicolumn{9}{c}{\textbf{Attribute Bias \& Structural Bias Comparison}} \\
\hline
        & \multicolumn{2}{c}{\textbf{Group 0 v.s. 1}} & & \multicolumn{2}{c}{\textbf{Group 0 v.s. 2}} & & \multicolumn{2}{c}{\textbf{Group 1 v.s. 2}} \\
\cline{2-3}  \cline{5-6}  \cline{8-9}
        & $b_{\text{attr}}$       & $b_{\text{stru}}$   &   & $b_{\text{attr}}$       & $b_{\text{stru}}$   &   & $b_{\text{attr}}$       & $b_{\text{stru}}$      \\
        \hline
\textbf{Vanilla} & 13.7      & 25.5    &     & 26.5    & 48.8    &      & 11.0        & 20.4                \\
\textbf{EDITS}   & \textbf{5.33}      & \textbf{9.63}    &     & \textbf{13.4}    & \textbf{24.1}    &      & \textbf{4.73}        & \textbf{8.73}               \\
\hline
\hline
\multicolumn{9}{c}{\textbf{Utility Comparison}} \\
\hline
        & \multicolumn{4}{c}{\textbf{AUC}}  & \multicolumn{4}{c}{\textbf{F1}} \\
        \hline
\textbf{Vanilla} & \multicolumn{4}{c}{\textbf{67.09 $\pm$ 0.3\%}}  & \multicolumn{4}{c}{\textbf{64.50 $\pm$ 0.6\%}} \\
\textbf{EDITS}   & \multicolumn{4}{c}{67.05 $\pm$ 0.2\%}  & \multicolumn{4}{c}{62.91 $\pm$ 0.8\%} \\
\hline
\end{tabular}
\vspace{-1.7em}
\end{table}

% \begin{table}[]
% % \vspace{-2mm}
% \small
% \centering
% \caption{xxx.}
% \vspace{-3mm}
% \label{bias_results_extension}
% \begin{tabular}{cccccc}
% \hline
%       & \multicolumn{2}{c}{\textbf{Attribute Bias}} &  & \multicolumn{2}{c}{\textbf{Structural Bias}} \\
%       \cline{2-3}  \cline{5-6}
%       & \textbf{Vanilla}   & \textbf{EDITS}   &  & \textbf{Vanilla}   & \textbf{EDITS}  \\
%       \hline
% \textbf{Pokec} & x.xx             &\textbf{x.xx} ($-x.xx\%$)             &  &x.xx  & \textbf{x.xx} ($-x.xx\%$)           \\
% \hline
% \textbf{Oklahoma97} & x.xx             &\textbf{x.xx} ($-x.xx\%$)             &  &x.xx & \textbf{x.xx}  ($-x.xx\%$)          \\
% \hline
% \textbf{UNC28}   & x.xx       &\textbf{x.xx} ($-x.xx\%$)             &  &x.xx & \textbf{x.xx}  ($-x.xx\%$)          \\
% \hline
% \textbf{Facebook}   & x.xx       &\textbf{x.xx} ($-x.xx\%$)             &  &x.xx & \textbf{x.xx}  ($-x.xx\%$)          \\
% \hline
% \end{tabular}
% \end{table}

\end{document}

%% file: images/auc1.tikz
\definecolor{mycolor1}{rgb}{1,0.65,0}%
\definecolor{mycolor2}{rgb}{1,0.94,0}%
\definecolor{mycolor3}{rgb}{1.0, 0.25, 0.25}%
\definecolor{mycolor4}{rgb}{0.39, 0.58, 0.93}
\definecolor{mycolor5}{rgb}{0., 0.18, 0.39}

\pgfplotsset{compat=1.11,
        /pgfplots/ybar legend/.style={
        /pgfplots/legend image code/.code={%
        %\draw[##1,/tikz/.cd,yshift=-0.25em]
                %(0cm,0cm) rectangle (3pt,0.8em);},
        \draw[##1,/tikz/.cd,bar width=3pt,yshift=-0.2em,bar shift=0pt]
                plot coordinates {(0cm,0.8em)};},
},
}

\begin{tikzpicture}
\centering
\begin{axis}
[
bar width=3.5pt,
    width= 0.6\figurewidth,
    height=1.1\figureheight,
    ybar, % ybar command displays the graph in horizontal form, while the xbar command displays the graph in vertical form.
    ymin=20,
    ymax=80,
    enlarge x limits={0.23}, % these limits are used to shrink or expand the graph. The lesser the limit, the higher the graph will expand or grow. The greater the limit, the more graph will shrink. 
    grid, % --added
    xticklabels={ , , },
xtick={1,2,3},
extra x ticks={1,2,3},
xmin=1,
xmax=3,
extra x tick labels={Pokec-z,Pokec-n,UCSD34,German,Credit,Recid.},  % ,w/o-AD,w/o-SD,Vanilla
every extra x tick/.style={tick label style={fill=none, rotate=25,anchor=north}},
    grid style={line width=.15pt, draw=gray!35}, % --added, dashed, 
    legend style={at={(0.35, 0.95)}, % these are the measures of the bottom row containing surplus (wheat, Tea, rice), where -0.25 is the gap between the bottom row and the graph. 
      anchor=north, legend columns=-1}, % here, north is the position of the bottom legend row. You can specify the east, west, or south direction to shift the location. 
    ylabel={AUC}, % there should be no line gap between the rows here. Otherwise, latex will show an error.
    legend style={nodes={scale=0.8, transform shape}, /tikz/every even column/.append style={column sep=0.1cm}},
    % xtick style={
    %         /pgfplots/major tick length=0pt,
    %     }
    % symbolic x coords={German, Credit, Recidivism}, xtick=data
]%, nodes near coords, nodes near coords align={vertical}, ]

            \addplot+ [
            %  draw=mycolor1, fill=mycolor1,
            error bars/.cd,
                y dir=both,
                y explicit ,
                % error bar style={color=mycolor1},
        ] coordinates {
            (1, 67.83)+-(German,0.7)
            (2, 63.24)+-(Credit,0.5)
            (3, 63.33)+-(Credit,0.3)
            (4, 74.46)+-(German,0.7)
            (5, 73.62)+-(Credit,0.3)
            (6, 86.91)+-(Recidivism,0.4)
        }; %\addlegendentry{Vanilla}
        \label{vanilla}

        \addplot+ [
            error bars/.cd,
                y dir=both,
                y explicit ,
        ] coordinates {
            (1, 69.50)+-(German,0.4)
            (2, 65.20)+-(Credit,0.4)
            (3, 65.98)+-(Credit,0.5)
            (4, 74.73)+-(German,0.8)
            (5, 68.50)+-(Credit,1.2)
            (6, 87.85)+-(Recidivism,1.2)
        }; %\addlegendentry{FairGNN}
        \label{fairgnn}
        
                \addplot+ [
            error bars/.cd,
                y dir=both,
                y explicit ,
        ] coordinates {
            (1, 70.36)+-(German,0.6)
            (2, 66.23)+-(Credit,0.2)
            (3, 62.08)+-(Credit,1.6)
            (4, 66.94)+-(A,3.6)
            (5, 69.09)+-(B,0.1)
            (6, 80.56)+-(C,0.2)
        }; %\addlegendentry{NIFTY}
        \label{nifty}
        
        \addplot+ [
            error bars/.cd,
                y dir=both,
                y explicit ,
        ] coordinates {
            (1, 67.38)+-(German,0.3)
            (2, 61.82)+-(Credit,0.9)
            (3, 62.43)+-(Credit,0.9)
            (4, 71.01)+-(A,1.3)
            (5, 70.16)+-(B,0.6)
            (6, 85.96)+-(C,0.3)
        }; %\addlegendentry{EDITS}
        \label{adorn}
% \legend{FairGNN, NIFTY-GNN, ADORN}
\end{axis}
\end{tikzpicture}

%% file: images/auc2.tikz
\definecolor{mycolor1}{rgb}{1,0.65,0}%
\definecolor{mycolor2}{rgb}{1,0.94,0}%
\definecolor{mycolor3}{rgb}{1.0, 0.25, 0.25}%
\definecolor{mycolor4}{rgb}{0.39, 0.58, 0.93}
\definecolor{mycolor5}{rgb}{0., 0.18, 0.39}

\pgfplotsset{compat=1.11,
        /pgfplots/ybar legend/.style={
        /pgfplots/legend image code/.code={%
        %\draw[##1,/tikz/.cd,yshift=-0.25em]
                %(0cm,0cm) rectangle (3pt,0.8em);},
        \draw[##1,/tikz/.cd,bar width=3pt,yshift=-0.2em,bar shift=0pt]
                plot coordinates {(0cm,0.8em)};},
},
}

\begin{tikzpicture}
\centering
\begin{axis}
[
bar width=3.5pt,
    width= 0.6\figurewidth,
    height=1.1\figureheight,
    ybar, % ybar command displays the graph in horizontal form, while the xbar command displays the graph in vertical form.
    ymin=20,
    ymax=95,
    enlarge x limits={0.23}, % these limits are used to shrink or expand the graph. The lesser the limit, the higher the graph will expand or grow. The greater the limit, the more graph will shrink. 
    grid, % --added
    xticklabels={ , , },
xtick={1,2,3,4,5,6},
extra x ticks={1,2,3,4,5,6},
xmin=4,
xmax=6,
extra x tick labels={Pokec-z,Pokec-n,UCSD34,German,Credit,Recid.},  % ,w/o-AD,w/o-SD,Vanilla
every extra x tick/.style={tick label style={fill=none, rotate=25,anchor=north}},
    grid style={line width=.15pt, draw=gray!35}, % --added, dashed, 
    legend style={at={(0.35, 0.95)}, % these are the measures of the bottom row containing surplus (wheat, Tea, rice), where -0.25 is the gap between the bottom row and the graph. 
      anchor=north, legend columns=-1}, % here, north is the position of the bottom legend row. You can specify the east, west, or south direction to shift the location. 
    ylabel={AUC}, % there should be no line gap between the rows here. Otherwise, latex will show an error.
    legend style={nodes={scale=0.8, transform shape}, /tikz/every even column/.append style={column sep=0.1cm}},
    % xtick style={
    %         /pgfplots/major tick length=0pt,
    %     }
    % symbolic x coords={German, Credit, Recidivism}, xtick=data
]%, nodes near coords, nodes near coords align={vertical}, ]

            \addplot+ [
            %  draw=mycolor1, fill=mycolor1,
            error bars/.cd,
                y dir=both,
                y explicit ,
                % error bar style={color=mycolor1},
        ] coordinates {
            (1, 67.83)+-(German,0.7)
            (2, 63.24)+-(Credit,0.5)
            (3, 63.33)+-(Credit,0.3)
            (4, 74.46)+-(German,0.7)
            (5, 73.62)+-(Credit,0.3)
            (6, 86.91)+-(Recidivism,0.4)
        }; %\addlegendentry{Vanilla}
        \label{vanilla}

        \addplot+ [
            error bars/.cd,
                y dir=both,
                y explicit ,
        ] coordinates {
            (1, 69.50)+-(German,0.4)
            (2, 65.20)+-(Credit,0.4)
            (3, 65.98)+-(Credit,0.5)
            (4, 74.73)+-(German,0.8)
            (5, 68.50)+-(Credit,1.2)
            (6, 87.85)+-(Recidivism,1.2)
        }; %\addlegendentry{FairGNN}
        \label{fairgnn}
        
                \addplot+ [
            error bars/.cd,
                y dir=both,
                y explicit ,
        ] coordinates {
            (1, 70.36)+-(German,0.6)
            (2, 66.23)+-(Credit,0.2)
            (3, 62.08)+-(Credit,1.6)
            (4, 66.94)+-(A,3.6)
            (5, 69.09)+-(B,0.1)
            (6, 80.56)+-(C,0.2)
        }; %\addlegendentry{NIFTY}
        \label{nifty}
        
        \addplot+ [
            error bars/.cd,
                y dir=both,
                y explicit ,
        ] coordinates {
            (1, 67.38)+-(German,0.3)
            (2, 61.82)+-(Credit,0.9)
            (3, 62.43)+-(Credit,0.9)
            (4, 71.01)+-(A,1.3)
            (5, 70.16)+-(B,0.6)
            (6, 85.96)+-(C,0.3)
        }; %\addlegendentry{EDITS}
        \label{adorn}
% \legend{FairGNN, NIFTY-GNN, ADORN}
\end{axis}
\end{tikzpicture}

%% file: images/sp1.tikz
\definecolor{mycolor1}{rgb}{1,0.65,0}%
\definecolor{mycolor2}{rgb}{1,0.94,0}%
\definecolor{mycolor3}{rgb}{1.0, 0.25, 0.25}%
\definecolor{mycolor4}{rgb}{0.39, 0.58, 0.93}
\definecolor{mycolor5}{rgb}{0., 0.18, 0.39}

\pgfplotsset{compat=1.11,
        /pgfplots/ybar legend/.style={
        /pgfplots/legend image code/.code={%
        %\draw[##1,/tikz/.cd,yshift=-0.25em]
                %(0cm,0cm) rectangle (3pt,0.8em);},
        \draw[##1,/tikz/.cd,bar width=3pt,yshift=-0.2em,bar shift=0pt]
                plot coordinates {(0cm,0.8em)};},
},
}

\begin{tikzpicture}
\centering
\begin{axis}
[
bar width=3.5pt,
    width= 0.6\figurewidth,
    height=1.1\figureheight,
    ybar, % ybar command displays the graph in horizontal form, while the xbar command displays the graph in vertical form.
    ymin=0,
    ymax=8,
    enlarge x limits={0.23}, % these limits are used to shrink or expand the graph. The lesser the limit, the higher the graph will expand or grow. The greater the limit, the more graph will shrink. 
    grid, % --added
    xticklabels={ , , },
xtick={1,2,3},
extra x ticks={1,2,3},
xmin=1,
xmax=3,
extra x tick labels={Pokec-z,Pokec-n,UCSD34,German,Credit,Recid.},  % ,w/o-AD,w/o-SD,Vanilla
every extra x tick/.style={tick label style={fill=none, rotate=25,anchor=north}},
    grid style={line width=.15pt, draw=gray!35}, % --added, dashed, 
    legend style={at={(0.35, 0.95)}, % these are the measures of the bottom row containing surplus (wheat, Tea, rice), where -0.25 is the gap between the bottom row and the graph. 
      anchor=north, legend columns=-1}, % here, north is the position of the bottom legend row. You can specify the east, west, or south direction to shift the location. 
    ylabel={$\Delta_{SP}$}, % there should be no line gap between the rows here. Otherwise, latex will show an error.
    legend style={nodes={scale=0.8, transform shape}, /tikz/every even column/.append style={column sep=0.1cm}},
    % xtick style={
    %         /pgfplots/major tick length=0pt,
    %     }
    % symbolic x coords={German, Credit, Recidivism}, xtick=data
]%, nodes near coords, nodes near coords align={vertical}, ]

            \addplot+ [
            %  draw=mycolor1, fill=mycolor1,
            error bars/.cd,
                y dir=both,
                y explicit ,
                % error bar style={color=mycolor1},
        ] coordinates {
            (1,5.7)+-(Credit,1.2)  % 5.7 1.2
            (2,3.36)+-(Recidivism,0.4)  % 3.36 0.4
            (3,1.27)+-(Recidivism,0.4)
            (4, 74.46)+-(German,0.7)
            (5, 73.62)+-(Credit,0.3)
            (6, 86.91)+-(Recidivism,0.4)
        }; %\addlegendentry{Vanilla}
        \label{vanilla}

        \addplot+ [
            error bars/.cd,
                y dir=both,
                y explicit ,
        ] coordinates {
            (1,3.29)+-(Credit,0.7)  % (1,3.29)+-(Credit,0.7)
            (2,1.58)+-(Recidivism,0.5)  % (2,1.58)+-(Recidivism,0.5)
            (3,0.43)+-(Recidivism,0.1)
            (4, 74.73)+-(German,0.8)
            (5, 68.50)+-(Credit,1.2)
            (6, 87.85)+-(Recidivism,1.2)
        }; %\addlegendentry{FairGNN}
        \label{fairgnn}
        
                \addplot+ [
            error bars/.cd,
                y dir=both,
                y explicit ,
        ] coordinates {
            (1,2.30)+-(Credit,0.5)
            (2,1.13)+-(Recidivism,0.3)
            (3,0.1)+-(Recidivism,0.05)
            (4, 66.94)+-(A,3.6)
            (5, 69.09)+-(B,0.1)
            (6, 80.56)+-(C,0.2)
        }; %\addlegendentry{NIFTY}
        \label{nifty}
        
        \addplot+ [
            error bars/.cd,
                y dir=both,
                y explicit ,
        ] coordinates {
            (1,2.74)+-(Credit,0.9)
            (2,0.91)+-(Recidivism,0.8)
            (3,0.27)+-(Recidivism,0.1)
            (4, 71.01)+-(A,1.3)
            (5, 70.16)+-(B,0.6)
            (6, 85.96)+-(C,0.3)
        }; %\addlegendentry{EDITS}
        \label{adorn}
% \legend{FairGNN, NIFTY-GNN, ADORN}
\end{axis}
\end{tikzpicture}

%% file: images/sp2.tikz
% \pgfplotsset{compat=1.16,
% /pgfplots/broken ybar legend/.style={
% /pgfplots/legend image code/.code={
% \draw [##1,/tikz/.cd,bar width=3pt,yshift=-0.2em,bar shift=0pt,yscale=2]
% plot coordinates {(0cm,0.8em) (2*\pgfplotbarwidth,0.6em)};
% \fill[white,decoration={zigzag,segment length=1pt,amplitude=0.3pt}]
% (-1.7pt,0.4em) -- (-1.7pt,0.5em) decorate{-- (1.7pt,0.5em)} --  (1.7pt,0.4em)
% decorate{-- (-1.7pt,0.4em)};
% },
% },}

\pgfplotsset{compat=1.11,
        /pgfplots/ybar legend/.style={
        /pgfplots/legend image code/.code={%
        %\draw[##1,/tikz/.cd,yshift=-0.25em]
                %(0cm,0cm) rectangle (3pt,0.8em);},
        \draw[##1,/tikz/.cd,bar width=3pt,yshift=-0.2em,bar shift=0pt]
                plot coordinates {(0cm,0.8em)};},
},
}

\begin{tikzpicture}
[pics/axis dicontinuity/.style={code={
    % \fill[white] (-0.21,-0.5) rectangle (0.21,0.5);
    % \draw (0,-0.6) -- (0,-0.4) -- ++ (0.2,0.2) -- ++(-0.4,0.4) 
    % -- ++ (0.2,0.2) -- (0,0.6);}},
    % \fill[white] (-0.05,-0.5) rectangle (0.05,0.5);
    % \draw (0,-0.6) -- (0,-0.4) -- ++ (0.2,0.2) -- ++(-0.4,0.4) 
    % -- ++ (0.2,0.2) -- (0,0.6);
    % \fill[white] (-0.05,-0.25) rectangle (0.05,0.25);
    % \draw (0,-0.3) -- (0,-0.1) -- ++ (0.08,0.1) -- ++(-0.16,0.2) 
    % -- ++ (0.08,0.06) -- (0,0.3);
    \fill[white] (-0.05,-0.3) rectangle (0.05,0.04);
    \draw (0,-0.3) -- ++ (0.08,0.1) -- ++(-0.16,0.16) 
     -- ++ (0.08,0.06) -- (0,0.3);
    }
    },
    pics/bar discontinuity/.style={code={
    }}]
    \begin{axis}[
    bar width=3.5pt,
        % legend pos=outer north east,
        width=0.6\figurewidth,
        height=1.1\figureheight,
        ybar,
        xticklabels={ , , },
xtick={1,2,3,4,5,6},
extra x ticks={1,2,3,4,5,6},
xmin=4,
xmax=6,
extra x tick labels={Pokec-z,Pokec-n,UCSD34,German,Credit,Recid.},  % ,w/o-AD,w/o-SD,Vanilla
every extra x tick/.style={tick label style={fill=none, rotate=25,anchor=north}},
        % symbolic x coords={German, Credit, Recidivism},
        % xtick={German, Credit, Recidivism},
        % xticklabels={German, Credit, Recidivism},
        grid, % --added
        grid style={line width=.15pt, draw=gray!35}, % --added, dashed, 
        ylabel=$\Delta_{SP}$,
        enlarge x limits={0.23},
        ytick={0,5,10,15,21.5,26.5,31.5},
        yticklabels={0,5,10,15,35,40,45},
        ymin=0,ymax=35,
        % ymajorgrids=true,
        legend columns = 2,
        scaled ticks=false,
        legend style={at={(0.95, 0.92)}, nodes={scale=0.8, transform shape}, /tikz/every even column/.append style={column sep=0.1cm}},
        % xtick style={
        %     /pgfplots/major tick length=0pt,
        % }
    ]
    
    \addplot+ [yscale=0.7,% broken ybar legend,
            error bars/.cd,
                % y dir=plus,
                y dir=both,
                %ymode=log,
                y explicit ,
        ] coordinates {
            (1,8.143)+-(Credit,1.7)  % 5.7 1.2
            (2,4.8)+-(Recidivism,0.6)  % 3.36 0.4
            (3,1.27)+-(Recidivism,0.4)
            (4,43.14)+-(German,2.5)  % 
            (5,18.47)+-(Credit,0.143)  % 12.93  0.1
            (6,11.27)+-(Recidivism,0.43)  % 7.89  0.3
        }; \addlegendentry{Vanilla} % vanilla

    \addplot+ [yscale=0.7,% broken ybar legend,
            error bars/.cd,
                % y dir=plus,
                y dir=both,
                %ymode=log,
                y explicit ,
        ] coordinates {
            (1,4.7)+-(Credit,1.0)  % (1,3.29)+-(Credit,0.7)
            (2,2.26)+-(Recidivism,0.71)  % (2,1.58)+-(Recidivism,0.5)
            (3,0.43)+-(Recidivism,0.1)
            (4,38.66)+-(German,5.2)
            (5,17.8)+-(Credit,0.57)  % 12.47  0.4
            (6,9.7)+-(Recidivism,0.429)  % 6.79  0.3
        }; \addlegendentry{FairGNN} % fairgnn

\addplot+ [
            error bars/.cd,
                y dir=both,
                y explicit ,
        ] coordinates {
            (1,2.30)+-(Credit,0.5)
            (2,1.13)+-(Recidivism,0.3)
            (3,0.1)+-(Recidivism,0.05)
            (4,3.13)+-(German,0.5)
            (5,10.58)+-(Credit,0.8)
            (6,5.20)+-(Recidivism,0.2)
        }; \addlegendentry{NIFTY}  % nifty
        
\addplot+ [     error bars/.cd,
                y dir=both,
                y explicit ,
        ] coordinates {
            (1,2.74)+-(Credit,0.9)
            (2,0.91)+-(Recidivism,0.8)
            (3,0.27)+-(Recidivism,0.1)
            (4,2.04)+-(German,0.4)
            (5,9.13)+-(Credit,1.2)
            (6,5.39)+-(Recidivism,0.2)
        }; \addlegendentry{EDITS} % adorn
    \path (axis description cs:0,0.6) coordinate (L) (axis description cs:1,0.6) coordinate (R);  
        % \legend{vanilla, fairgnn,nifty,adorn}
    \end{axis}
    \path (L) pic {axis dicontinuity} (R) pic {axis dicontinuity};
    \fill[white,decoration={zigzag,segment length=1.6mm,amplitude=0.3mm}]
    ([xshift=4.55mm,yshift=-2.5mm]L) -- ++ (0mm,1.5mm)
    decorate{-- ([xshift=-29.0mm,yshift=-0.8mm]R)}
    -- ++(-0mm,-2mm) decorate{-- cycle};
\end{tikzpicture}

%% file: images/to_bar_attr.tikz
\pgfplotsset{compat=1.11,
        /pgfplots/ybar legend/.style={
        /pgfplots/legend image code/.code={%
        %\draw[##1,/tikz/.cd,yshift=-0.25em]
                %(0cm,0cm) rectangle (3pt,0.8em);},
        \draw[##1,/tikz/.cd,bar width=3pt,yshift=-0.2em,bar shift=0pt]
                plot coordinates {(0cm,0.8em)};},
},
}

\begin{tikzpicture}
[pics/axis dicontinuity/.style={code={
    % \fill[white] (-0.21,-0.5) rectangle (0.21,0.5);
    % \draw (0,-0.6) -- (0,-0.4) -- ++ (0.2,0.2) -- ++(-0.4,0.4) 
    % -- ++ (0.2,0.2) -- (0,0.6);}},
    % \fill[white] (-0.05,-0.5) rectangle (0.05,0.5);
    % \draw (0,-0.6) -- (0,-0.4) -- ++ (0.2,0.2) -- ++(-0.4,0.4) 
    % -- ++ (0.2,0.2) -- (0,0.6);
    \fill[white] (-0.05,-0.25) rectangle (0.05,0.25);
    \draw (0,-0.3) -- (0,-0.1) -- ++ (0.08,0.1) -- ++(-0.16,0.2) 
    -- ++ (0.08,0.06) -- (0,0.3);
    }
    },
pics/bar discontinuity/.style={code={
    }}]
\begin{axis}[%
        ybar,
bar width=2.5pt,
enlarge x limits={0.25}, 
width=\figurewidth,
height=0.8\figureheight,
at={(0\figurewidth,0\figureheight)},
scale only axis,
legend style={at={(0.22, 0.20)}, anchor=south, legend columns=1}, 
xticklabels={ , , , },
xtick={1,2,3,4},
extra x ticks={1,2,3,4},
xmin=1,
xmax=4,
extra x tick labels={EDITS,*w/o-SD,*w/o-AD,Vanilla},  % ,w/o-AD,w/o-SD,Vanilla
every extra x tick/.style={tick label style={fill=none, rotate=25,anchor=north}},
grid, % --added
grid style={line width=.15pt, draw=gray!15}, % --added, dashed, 
ylabel={$b_{\text{attr}} (\times 10^{-3})$},
ytick={0,3,6,9,13.5,16.5},
yticklabels={0,3,6,9,15,18},
ymin=0,
ymax=16.5,
axis background/.style={fill=white},
axis background/.style={fill=white} % --added
]
    
%     \addplot+ [color=mycolor1, mark=otimes*, mark options={solid, mycolor1}, thick]
%   table[row sep=crcr]{%
% 1	2.38\\
% 2	2.63\\
% 3	6.33\\
% 4	6.33\\
% }; \label{german}
% % \addlegendentry{ATEX}

% \addplot+ [color=mycolor5, mark=triangle*, mark options={solid, mycolor5}, thick]
%   table[row sep=crcr]{%
% 1	0.68\\
% 2	1.02\\
% 3	1.78\\
% 4	1.78\\
% }; \label{credit}
% % \addlegendentry{Baseline}

% \addplot+ [color=mycolor3, mark=diamond*, mark options={solid, mycolor3}, thick]
%   table[row sep=crcr]{%
% 1	1.27\\  
% 2	2.51\\
% 3	15.39\\  % 17.1
% 4	15.39\\  % 17.1
% }; \label{biased x}

% \addplot+ [color=mycolor4, mark=square*, mark options={solid, mycolor4}, thick]
%   table[row sep=crcr]{%
% 1	2.5\\ 
% 2	2.5\\ 
% 3	7.98\\
% 4	7.98\\
% };\label{biased a}
    
    \addplot+ [yscale=1,% broken ybar legend,
            error bars/.cd,
                % y dir=plus,
                y dir=both,
                %ymode=log,
                y explicit ,
        ] coordinates {
            (1,2.38)
            (2,2.63)
            (3,6.33)
            (4,6.33)
        }; \addlegendentry{German} \label{german}

    \addplot+ [yscale=1,% broken ybar legend,
            error bars/.cd,
                % y dir=plus,
                y dir=both,
                %ymode=log,
                y explicit ,
        ] coordinates {
            (1,0.56)
            (2,1.02)
            (3,2.46)
            (4,2.46)
        }; \addlegendentry{Credit}  \label{credit}

\addplot+ [
            error bars/.cd,
                y dir=both,
                y explicit ,
        ] coordinates {
            (1,1.27)
            (2,2.51)
            (3,15.39)
            (4,15.39)
        }; \addlegendentry{Biased $\mathbf{X}$}  \label{biased x}
        
\addplot+ [     error bars/.cd,
                y dir=both,
                y explicit ,
        ] coordinates {
            (1,2.5)
            (2,2.5)
            (3,7.98)
            (4,7.98)
        }; \addlegendentry{Biased $\mathbf{A}$}  \label{biased a}
    \path (axis description cs:0,0.6) coordinate (L) (axis description cs:1,0.6) coordinate (R);  
        % \legend{vanilla, fairgnn,nifty,adorn}
    \end{axis}
    \path (L) pic {axis dicontinuity} (R) pic {axis dicontinuity};
    \fill[white,decoration={zigzag,segment length=0.8mm,amplitude=0.4mm}]
    ([xshift=16.5mm,yshift=-2mm]L) -- ++ (0,3mm)
    decorate{-- ([xshift=-2.5mm,yshift=2mm]R)}
    -- ++(0,-3mm) decorate{-- cycle};
\end{tikzpicture}

%% file: images/to_bar_stru.tikz
\pgfplotsset{compat=1.11,
        /pgfplots/ybar legend/.style={
        /pgfplots/legend image code/.code={%
        %\draw[##1,/tikz/.cd,yshift=-0.25em]
                %(0cm,0cm) rectangle (3pt,0.8em);},
        \draw[##1,/tikz/.cd,bar width=3pt,yshift=-0.2em,bar shift=0pt]
                plot coordinates {(0cm,0.8em)};},
},
}

\begin{tikzpicture}
[pics/axis dicontinuity/.style={code={
    % \fill[white] (-0.21,-0.5) rectangle (0.21,0.5);
    % \draw (0,-0.6) -- (0,-0.4) -- ++ (0.2,0.2) -- ++(-0.4,0.4) 
    % -- ++ (0.2,0.2) -- (0,0.6);}},
    % \fill[white] (-0.05,-0.5) rectangle (0.05,0.5);
    % \draw (0,-0.6) -- (0,-0.4) -- ++ (0.2,0.2) -- ++(-0.4,0.4) 
    % -- ++ (0.2,0.2) -- (0,0.6);
    \fill[white] (-0.05,-0.25) rectangle (0.05,0.25);
    \draw (0,-0.3) -- (0,-0.1) -- ++ (0.08,0.1) -- ++(-0.16,0.2) 
    -- ++ (0.08,0.06) -- (0,0.3);
    }
    },
pics/bar discontinuity/.style={code={
    }}]
\begin{axis}[%
        ybar,
bar width=2.5pt,
enlarge x limits={0.215}, 
width=\figurewidth,
height=0.8\figureheight,
at={(0\figurewidth,0\figureheight)},
scale only axis,
xticklabels={ , , , },
xtick={1,2,3,4},
extra x ticks={1,2,3,4},
extra x tick labels={EDITS,*w/o-SD,*w/o-AD,Vanilla},  % ,w/o-AD,w/o-SD,Vanilla
every extra x tick/.style={tick label style={fill=none, rotate=25,anchor=north}},
grid, % --added
grid style={line width=.15pt, draw=gray!15}, % --added, dashed, 
ylabel={$b_{\text{stru}} (\times 10^{-3})$},
ytick={0,4,8,12,18,21.6},
yticklabels={0,4,8,12,20,24},
ymin=0,
ymax=23,
axis background/.style={fill=white},
axis background/.style={fill=white} % --added
]
    
% \addplot [color=mycolor1, mark=otimes*, mark options={solid, mycolor1}, thick]
%   table[row sep=crcr]{%
% 1	3.54\\
% 2	4.78\\
% 3	8.88\\
% 4	10.4\\
% }; \label{german}
% % \addlegendentry{ATEX}

% \addplot [color=mycolor5, mark=triangle*, mark options={solid, mycolor5}, thick]
%   table[row sep=crcr]{%
% 1	1.74\\
% 2	2.24\\
% 3	2.95\\
% 4	3.34\\
% }; \label{credit}
% % \addlegendentry{Baseline}

%     \path (axis description cs:0,0.6) coordinate (L) (axis description cs:1,0.6) coordinate (R);  
%         % \legend{vanilla, fairgnn,nifty,adorn}

% \addplot [color=mycolor3, mark=diamond*, mark options={solid, mycolor3}, thick]
%   table[row sep=crcr]{%
% 1	1.97\\  
% 2	3.79\\
% 3	17.64\\ % 19.6
% 4	21.0\\ % 23.3
% }; \label{biased x}

% \addplot [color=mycolor4, mark=square*, mark options={solid, mycolor4}, thick]
%   table[row sep=crcr]{%
% 1	3.4\\ 
% 2	5.0\\ 
% 3	11.9\\
% 4	12.4\\
% };\label{biased a}
    
    \addplot+ [yscale=1,% broken ybar legend,
            error bars/.cd,
                % y dir=plus,
                y dir=both,
                %ymode=log,
                y explicit ,
        ] coordinates {
            (1,3.54)
            (2,4.78)
            (3,8.88)
            (4,10.4)
        }; \label{german}

    \addplot+ [yscale=1,% broken ybar legend,
            error bars/.cd,
                % y dir=plus,
                y dir=both,
                %ymode=log,
                y explicit ,
        ] coordinates {
            (1,2.36)
            (2,2.64)
            (3,3.25)
            (4,4.45)
        }; \label{credit}

\addplot+ [
            error bars/.cd,
                y dir=both,
                y explicit ,
        ] coordinates {
            (1,1.97)
            (2,3.79)
            (3,17.64)
            (4,21.0)
        }; \label{biased x}
        
\addplot+ [     error bars/.cd,
                y dir=both,
                y explicit ,
        ] coordinates {
            (1,3.4)
            (2,5.0)
            (3,11.9)
            (4,12.4)
        }; \label{biased a}
    \path (axis description cs:0,0.6) coordinate (L) (axis description cs:1,0.6) coordinate (R);  
        % \legend{vanilla, fairgnn,nifty,adorn}
    \end{axis}
    \path (L) pic {axis dicontinuity} (R) pic {axis dicontinuity};
    \fill[white,decoration={zigzag,segment length=0.8mm,amplitude=0.4mm}]
    ([xshift=2.5mm,yshift=-2mm]L) -- ++ (0,3mm)
    decorate{-- ([xshift=-2.5mm,yshift=2mm]R)}
    -- ++(0,-3mm) decorate{-- cycle};
\end{tikzpicture}

%% file: images/to_bar_auc.tikz
\pgfplotsset{compat=1.11,
        /pgfplots/ybar legend/.style={
        /pgfplots/legend image code/.code={%
        %\draw[##1,/tikz/.cd,yshift=-0.25em]
                %(0cm,0cm) rectangle (3pt,0.8em);},
        \draw[##1,/tikz/.cd,bar width=3pt,yshift=-0.2em,bar shift=0pt]
                plot coordinates {(0cm,0.8em)};},
},
}

\begin{tikzpicture}
[pics/axis dicontinuity/.style={code={
    % \fill[white] (-0.21,-0.5) rectangle (0.21,0.5);
    % \draw (0,-0.6) -- (0,-0.4) -- ++ (0.2,0.2) -- ++(-0.4,0.4) 
    % -- ++ (0.2,0.2) -- (0,0.6);}},
    % \fill[white] (-0.05,-0.5) rectangle (0.05,0.5);
    % \draw (0,-0.6) -- (0,-0.4) -- ++ (0.2,0.2) -- ++(-0.4,0.4) 
    % -- ++ (0.2,0.2) -- (0,0.6);
    \fill[white] (-0.05,-0.25) rectangle (0.05,0.25);
    \draw (0,-0.3) -- (0,-0.1) -- ++ (0.08,0.1) -- ++(-0.16,0.2) 
    -- ++ (0.08,0.06) -- (0,0.3);
    }
    },
pics/bar discontinuity/.style={code={
    }}]
\begin{axis}[%
        ybar,
bar width=2.5pt,
enlarge x limits={0.215}, 
width=\figurewidth,
height=0.8\figureheight,
at={(0\figurewidth,0\figureheight)},
scale only axis,
xticklabels={ , , , },
xtick={1,2,3,4},
extra x ticks={1,2,3,4},
extra x tick labels={EDITS,*w/o-SD,*w/o-AD,Vanilla},  % ,w/o-AD,w/o-SD,Vanilla
every extra x tick/.style={tick label style={fill=none, rotate=25,anchor=north}},
legend style={at={(0.50, 0.12)}, anchor=south, legend columns=2}, 
grid, % --added
grid style={line width=.15pt, draw=gray!15}, % --added, dashed, 
ylabel={AUC},
ymin=20,
ymax=100,
axis background/.style={fill=white},
axis background/.style={fill=white} % --added
]
    
% \addplot [color=mycolor1, mark=otimes*, mark options={solid, mycolor1}, thick]
%   table[row sep=crcr]{%
% 1	70.26\\
% 2	72.20\\
% 3	69.52\\
% 4	74.46\\
% }; \label{german}
% \addlegendentry{German}

% \addplot [color=mycolor5, mark=triangle*, mark options={solid, mycolor5}, thick]
%   table[row sep=crcr]{%
% 1	70.16\\
% 2	72.71\\
% 3	73.83\\
% 4	73.62\\
% }; \label{credit}
% \addlegendentry{Credit}

% \addplot [color=mycolor3, mark=diamond*, mark options={solid, mycolor3}, thick]
%   table[row sep=crcr]{%
% 1	86.40\\  
% 2	93.30\\
% 3	85.70\\  
% 4	93.08\\ 
% }; \label{biased x}
% \addlegendentry{Biased $\mathbf{X}$}

% \addplot [color=mycolor4, mark=square*, mark options={solid, mycolor4}, thick]
%   table[row sep=crcr]{%
% 1	81.3\\ 
% 2	82.02\\ % waiting
% 3	83.2\\
% 4	83.69\\
% };\label{biased a}
% \addlegendentry{Biased $\mathbf{A}$}
    
    \addplot+ [yscale=1,% broken ybar legend,
            error bars/.cd,
                % y dir=plus,
                y dir=both,
                %ymode=log,
                y explicit ,
        ] coordinates {
            (1,71.01)
            (2,72.20)
            (3,69.52)
            (4,74.46)
        }; \label{german}

    \addplot+ [yscale=1,% broken ybar legend,
            error bars/.cd,
                % y dir=plus,
                y dir=both,
                %ymode=log,
                y explicit ,
        ] coordinates {
            (1,70.16)
            (2,72.71)
            (3,73.83)
            (4,73.62)
        }; \label{credit}

\addplot+ [
            error bars/.cd,
                y dir=both,
                y explicit ,
        ] coordinates {
            (1,86.40)
            (2,93.30)
            (3,85.70)
            (4,93.08)
        }; \label{biased x}
        
\addplot+ [     error bars/.cd,
                y dir=both,
                y explicit ,
        ] coordinates {
            (1,81.3)
            (2,82.02)
            (3,83.2)
            (4,83.69)
        }; \label{biased a}
    \path (axis description cs:0,0.6) coordinate (L) (axis description cs:1,0.6) coordinate (R);  
        % \legend{vanilla, fairgnn,nifty,adorn}
    \end{axis}
    % \path (L) pic {axis dicontinuity} (R) pic {axis dicontinuity};
    % \fill[white,decoration={zigzag,segment length=0.1mm,amplitude=0.4mm}]
    % ([xshift=1.5mm,yshift=-4mm]L) -- ++ (0,3mm)
    % decorate{-- ([xshift=-2.5mm,yshift=2mm]R)}
    % -- ++(0,-3mm) decorate{-- cycle};
\end{tikzpicture}

%% file: images/to_bar_sp.tikz
\pgfplotsset{compat=1.11,
        /pgfplots/ybar legend/.style={
        /pgfplots/legend image code/.code={%
        %\draw[##1,/tikz/.cd,yshift=-0.25em]
                %(0cm,0cm) rectangle (3pt,0.8em);},
        \draw[##1,/tikz/.cd,bar width=3pt,yshift=-0.2em,bar shift=0pt]
                plot coordinates {(0cm,0.8em)};},
},
}

\begin{tikzpicture}
[pics/axis dicontinuity/.style={code={
    % \fill[white] (-0.21,-0.5) rectangle (0.21,0.5);
    % \draw (0,-0.6) -- (0,-0.4) -- ++ (0.2,0.2) -- ++(-0.4,0.4) 
    % -- ++ (0.2,0.2) -- (0,0.6);}},
    % \fill[white] (-0.05,-0.5) rectangle (0.05,0.5);
    % \draw (0,-0.6) -- (0,-0.4) -- ++ (0.2,0.2) -- ++(-0.4,0.4) 
    % -- ++ (0.2,0.2) -- (0,0.6);
    \fill[white] (-0.05,-0.25) rectangle (0.05,0.25);
    \draw (0,-0.3) -- (0,-0.1) -- ++ (0.08,0.1) -- ++(-0.16,0.2) 
    -- ++ (0.08,0.06) -- (0,0.3);
    }
    },
pics/bar discontinuity/.style={code={
    }}]
\begin{axis}[%
        ybar,
bar width=2.5pt,
enlarge x limits={0.215}, 
width=\figurewidth,
height=0.8\figureheight,
at={(0\figurewidth,0\figureheight)},
scale only axis,
xticklabels={ , , , },
xtick={1,2,3,4},
extra x ticks={1,2,3,4},
extra x tick labels={EDITS,*w/o-SD,*w/o-AD,Vanilla},  % ,w/o-AD,w/o-SD,Vanilla
every extra x tick/.style={tick label style={fill=none, rotate=25,anchor=north}},
grid, % --added
grid style={line width=.15pt, draw=gray!15}, % --added, dashed, 
ylabel={$\Delta_{SP}$},
ytick={0,5,10,15,20,22.5},
yticklabels={0,5,10,15,40,45},
ymin=0,
ymax=26,
axis background/.style={fill=white},
axis background/.style={fill=white} % --added
]
    
% \addplot [color=mycolor1,mark=otimes*, mark options={solid, mycolor1}, thick]
%   table[row sep=crcr]{%
% 1	0.73\\
% 2	11.5\\
% 3	1.73\\
% 4	21.55\\  % 43.1
% }; \label{german}
% % \addlegendentry{ATEX}

% \addplot [color=mycolor5, mark=triangle*, mark options={solid, mycolor5}, thick]
%   table[row sep=crcr]{%
% 1	9.13\\
% 2	12.1\\
% 3	10.4\\
% 4	12.9\\
% }; \label{credit}
% % \addlegendentry{Baseline}

%     \path (axis description cs:0,0.6) coordinate (L) (axis description cs:1,0.6) coordinate (R);  

% \addplot [color=mycolor3, mark=diamond*, mark options={solid, mycolor3}, thick]
%   table[row sep=crcr]{%
% 1	0.8\\  
% 2	4.2\\
% 3	0.8\\ 
% 4	5.8\\ 
% }; \label{biased x}

% \addplot [color=mycolor4, mark=square*, mark options={solid, mycolor4}, thick]
%   table[row sep=crcr]{%
% 1	5.3\\ 
% 2	11.1\\ 
% 3	5.9\\
% 4	13.3\\
% };\label{biased a}
    
    \addplot+ [yscale=1,% broken ybar legend,
            error bars/.cd,
                % y dir=plus,
                y dir=both,
                %ymode=log,
                y explicit ,
        ] coordinates {
            (1,2.04)
            (2,11.5)
            (3,3.73)
            (4,21.55)
        }; \label{german}

    \addplot+ [yscale=1,% broken ybar legend,
            error bars/.cd,
                % y dir=plus,
                y dir=both,
                %ymode=log,
                y explicit ,
        ] coordinates {
            (1,9.13)
            (2,12.1)
            (3,10.4)
            (4,12.9)
        }; \label{credit}

\addplot+ [
            error bars/.cd,
                y dir=both,
                y explicit ,
        ] coordinates {
            (1,0.8)
            (2,4.2)
            (3,0.8)
            (4,5.8)
        }; \label{biased x}
        
\addplot+ [     error bars/.cd,
                y dir=both,
                y explicit ,
        ] coordinates {
            (1,5.3)
            (2,11.1)
            (3,5.9)
            (4,13.3)
        }; \label{biased a}
    \path (axis description cs:0,0.6) coordinate (L) (axis description cs:1,0.6) coordinate (R);  
        % \legend{vanilla, fairgnn,nifty,adorn}
    \end{axis}
    \path (L) pic {axis dicontinuity} (R) pic {axis dicontinuity};
    \fill[white,decoration={zigzag,segment length=0.8mm,amplitude=0.4mm}]
    ([xshift=2.5mm,yshift=-2mm]L) -- ++ (0,3mm)
    decorate{-- ([xshift=-2.5mm,yshift=2mm]R)}
    -- ++(0,-3mm) decorate{-- cycle};
\end{tikzpicture}

%% file: images/non_binary_sp.tikz
\definecolor{mycolor1}{rgb}{1,0.65,0}%
\definecolor{mycolor2}{rgb}{1,0.94,0}%
\definecolor{mycolor3}{rgb}{1.0, 0.25, 0.25}%
\definecolor{mycolor4}{rgb}{0.39, 0.58, 0.93}
\definecolor{mycolor5}{rgb}{0., 0.18, 0.39}

\pgfplotsset{compat=1.11,
        /pgfplots/ybar legend/.style={
        /pgfplots/legend image code/.code={%
        %\draw[##1,/tikz/.cd,yshift=-0.25em]
                %(0cm,0cm) rectangle (3pt,0.8em);},
        \draw[##1,/tikz/.cd,bar width=3pt,yshift=-0.2em,bar shift=0pt]
                plot coordinates {(0cm,0.8em)};},
},
}

\begin{tikzpicture}
\centering
\begin{axis}
[
bar width=6.5pt,
    width= 0.6\figurewidth,
    height=1.1\figureheight,
    ybar, % ybar command displays the graph in horizontal form, while the xbar command displays the graph in vertical form.
    ymin=0,
    ymax=22,
    enlarge x limits={0.25}, % these limits are used to shrink or expand the graph. The lesser the limit, the higher the graph will expand or grow. The greater the limit, the more graph will shrink. 
    grid, % --added
    xticklabels={ , , },
xtick={1,2,3},
extra x ticks={1,2,3},
xmin=1,
xmax=3,
extra x tick labels={ 0 vs. 1, 0 vs. 2, 1 vs. 2.},  % ,w/o-AD,w/o-SD,Vanilla
every extra x tick/.style={tick label style={fill=none, rotate=25,anchor=north}},
    grid style={line width=.15pt, draw=gray!35}, % --added, dashed, 
    legend style={at={(0.5, 0.95)}, % these are the measures of the bottom row containing surplus (wheat, Tea, rice), where -0.25 is the gap between the bottom row and the graph. 
      anchor=north, legend columns=1}, % here, north is the position of the bottom legend row. You can specify the east, west, or south direction to shift the location. 
    ylabel={$\Delta_{SP}$}, % there should be no line gap between the rows here. Otherwise, latex will show an error.
    legend style={nodes={scale=0.8, transform shape}, /tikz/every even column/.append style={column sep=0.1cm}},
    % symbolic x coords={German, Credit, Recidivism}, xtick=data
]%, nodes near coords, nodes near coords align={vertical}, ]

            \addplot+ [
            %  draw=mycolor1, fill=mycolor1,
            error bars/.cd,
                y dir=both,
                y explicit ,
                % error bar style={color=mycolor1},
        ] coordinates {
            (1, 11.22)+-(German,3.2)  % 0-1
            (2, 3.5)+-(Credit,2.5)  % 0-2
            (3, 14.73)+-(Recidivism,1.6)  % 1-2
        }; %\addlegendentry{Vanilla}
        \label{Vanilla}

        \addplot+ [
            error bars/.cd,
                y dir=both,
                y explicit ,
        ] coordinates {
            (1, 2.8)+-(German,2.3)  % 0-1
            (2, 2.2)+-(Credit,0.7)  % 0-2
            (3, 7.03)+-(Recidivism,1.4)  % 1-2
        }; %\addlegendentry{FairGNN}
        \label{EDITS}
        
\legend{Vanilla,EDITS}
\end{axis}
\end{tikzpicture}

% \definecolor{mycolor1}{rgb}{1,0.65,0}%
% \definecolor{mycolor2}{rgb}{1,0.94,0}%
% \definecolor{mycolor3}{rgb}{1.0, 0.25, 0.25}%
% \definecolor{mycolor4}{rgb}{0.39, 0.58, 0.93}
% \definecolor{mycolor5}{rgb}{0., 0.18, 0.39}

% %
% \begin{tikzpicture}

% \begin{axis}[%
% width=0.951\figurewidth,
% height=\figureheight,
% at={(0\figurewidth,0\figureheight)},
% scale only axis,
% scaled x ticks=true,
% xticklabels={1,2,3,4,5},
% xtick={1,2,3,4,5},
% xmin=1,
% xmax=5,
% xlabel style={font=\color{white!15!black}},
% xlabel={ATEX iterations},
% ymin=0.10,
% ymax=0.50,
% grid, % --added
% grid style={line width=.15pt, draw=gray!15}, % --added, dashed, 
% ylabel style={font=\color{white!15!black}},
% ylabel={SSIM},
% axis background/.style={fill=white},
% legend columns = 2,
% legend style={legend cell align=left, align=left, draw=white!15!black, nodes={scale=0.7}, at={(0.90, 0.97)}},
% axis background/.style={fill=white} % --added
% ]

% \addplot [color=mycolor5, mark=square*, mark options={solid, mycolor5}, thick]
%   table[row sep=crcr]{%
% 1	0.310\\
% 2	0.321\\
% 3	0.356\\
% 4	0.354\\
% 5	0.354\\
% };
% \addlegendentry{ATEX}

% \addplot [color=mycolor5, mark=triangle, mark options={solid, mycolor5}, thick]
%   table[row sep=crcr]{%
% 1	0.186\\
% 2	0.185\\
% 3	0.185\\
% 4	0.185\\
% 5	0.185\\
% };
% \addlegendentry{Baseline}

% \end{axis}

% \end{tikzpicture}%

%% file: images/non_binary_eo.tikz
\definecolor{mycolor1}{rgb}{1,0.65,0}%
\definecolor{mycolor2}{rgb}{1,0.94,0}%
\definecolor{mycolor3}{rgb}{1.0, 0.25, 0.25}%
\definecolor{mycolor4}{rgb}{0.39, 0.58, 0.93}
\definecolor{mycolor5}{rgb}{0., 0.18, 0.39}

\pgfplotsset{compat=1.11,
        /pgfplots/ybar legend/.style={
        /pgfplots/legend image code/.code={%
        %\draw[##1,/tikz/.cd,yshift=-0.25em]
                %(0cm,0cm) rectangle (3pt,0.8em);},
        \draw[##1,/tikz/.cd,bar width=3pt,yshift=-0.2em,bar shift=0pt]
                plot coordinates {(0cm,0.8em)};},
},
}

\begin{tikzpicture}
\centering
\begin{axis}
[
bar width=6.5pt,
    width= 0.6\figurewidth,
    height=1.1\figureheight,
    ybar, % ybar command displays the graph in horizontal form, while the xbar command displays the graph in vertical form.
    ymin=0,
    ymax=17,
    enlarge x limits={0.25}, % these limits are used to shrink or expand the graph. The lesser the limit, the higher the graph will expand or grow. The greater the limit, the more graph will shrink. 
    grid, % --added
    xticklabels={ , , },
xtick={1,2,3},
extra x ticks={1,2,3},
xmin=1,
xmax=3,
extra x tick labels={ 0 vs. 1, 0 vs. 2, 1 vs. 2.},  % ,w/o-AD,w/o-SD,Vanilla
every extra x tick/.style={tick label style={fill=none, rotate=25,anchor=north}},
    grid style={line width=.15pt, draw=gray!35}, % --added, dashed, 
    legend style={at={(0.35, 0.95)}, % these are the measures of the bottom row containing surplus (wheat, Tea, rice), where -0.25 is the gap between the bottom row and the graph. 
      anchor=north, legend columns=-1}, % here, north is the position of the bottom legend row. You can specify the east, west, or south direction to shift the location. 
    ylabel={$\Delta_{EO}$}, % there should be no line gap between the rows here. Otherwise, latex will show an error.
    legend style={nodes={scale=0.8, transform shape}, /tikz/every even column/.append style={column sep=0.1cm}},
    % symbolic x coords={German, Credit, Recidivism}, xtick=data
]%, nodes near coords, nodes near coords align={vertical}, ]

            \addplot+ [
            %  draw=mycolor1, fill=mycolor1,
            error bars/.cd,
                y dir=both,
                y explicit ,
                % error bar style={color=mycolor1},
        ] coordinates {
            (1, 8.446)+-(German,3.1)
            (2, 6.5)+-(Credit,3.7)
            (3, 4.04)+-(Recidivism,1.2)
        }; %\addlegendentry{Vanilla}
        \label{vanilla}

        \addplot+ [
            error bars/.cd,
                y dir=both,
                y explicit ,
        ] coordinates {
            (1,4.0)+-(German,0.9)
            (2, 1.6)+-(Credit,1.0)
            (3, 3.4)+-(Recidivism,0.95)
        }; %\addlegendentry{FairGNN}
        \label{fairgnn}
        
% \legend{FairGNN, NIFTY-GNN, ADORN}
\end{axis}
\end{tikzpicture}

% \definecolor{mycolor1}{rgb}{1,0.65,0}%
% \definecolor{mycolor2}{rgb}{1,0.94,0}%
% \definecolor{mycolor3}{rgb}{1.0, 0.25, 0.25}%
% \definecolor{mycolor4}{rgb}{0.39, 0.58, 0.93}
% \definecolor{mycolor5}{rgb}{0., 0.18, 0.39}

% %
% \begin{tikzpicture}

% \begin{axis}[%
% width=0.951\figurewidth,
% height=\figureheight,
% at={(0\figurewidth,0\figureheight)},
% scale only axis,
% scaled x ticks=true,
% xticklabels={1,2,3,4,5},
% xtick={1,2,3,4,5},
% xmin=1,
% xmax=5,
% xlabel style={font=\color{white!15!black}},
% xlabel={ATEX iterations},
% ymin=0.10,
% ymax=0.50,
% grid, % --added
% grid style={line width=.15pt, draw=gray!15}, % --added, dashed, 
% ylabel style={font=\color{white!15!black}},
% ylabel={SSIM},
% axis background/.style={fill=white},
% legend columns = 2,
% legend style={legend cell align=left, align=left, draw=white!15!black, nodes={scale=0.7}, at={(0.90, 0.97)}},
% axis background/.style={fill=white} % --added
% ]

% \addplot [color=mycolor5, mark=square*, mark options={solid, mycolor5}, thick]
%   table[row sep=crcr]{%
% 1	0.310\\
% 2	0.321\\
% 3	0.356\\
% 4	0.354\\
% 5	0.354\\
% };
% \addlegendentry{ATEX}

% \addplot [color=mycolor5, mark=triangle, mark options={solid, mycolor5}, thick]
%   table[row sep=crcr]{%
% 1	0.186\\
% 2	0.185\\
% 3	0.185\\
% 4	0.185\\
% 5	0.185\\
% };
% \addlegendentry{Baseline}

% \end{axis}

% \end{tikzpicture}%